\PassOptionsToPackage{table}{xcolor}
\documentclass[acmtog,screen]{acmart}

\usepackage{booktabs} 

\usepackage[ruled]{algorithm2e} 

\SetAlFnt{\smallhttps://www.overleaf.com/project/644a114e232156b371bd5ca1}
\SetAlCapFnt{\small}
\SetAlCapNameFnt{\small}
\SetAlCapHSkip{0pt}

\setcopyright{iw3c2w3}
\acmJournal{TOG}
\acmYear{2023} \acmVolume{42} \acmNumber{6} \acmArticle{198} \acmMonth{12} \acmPrice{15.00}\acmDOI{10.1145/3618319}

\usepackage{amsfonts}
\usepackage{pifont}
\usepackage{multirow}

\begin{document}
	\title{Towards Garment Sewing Pattern Reconstruction from a Single Image}
	\author{Lijuan Liu}
	\affiliation{
		\institution{Sea AI Lab}
		\country{Singapore}}
	\email{liulj@sea.com}
        \authornote{Joint first authors.}
	\author{Xiangyu Xu}
	\affiliation{%
		  \institution{Xi'an Jiaotong University}
		 \country{China}}
        \affiliation{
            \institution{Sea AI Lab}
            \country{Singapore}
        }
	\email{xuxiangyu2014@gmail.com}
        \authornotemark[1]
        \authornote{Work done at Sea AI Lab.}
	\author{Zhijie Lin}
	\affiliation{%
		 \institution{Sea AI Lab}
		 \country{Singapore}
            }
	\email{jiewinchester@gmail.com}
        \authornotemark[1]
	\author{Jiabin Liang}
	\affiliation{%
		  \institution{Sea AI Lab}
		 \country{Singapore}
		}
	\email{ liangjb@sea.com}
        \authornotemark[1]
 
	\author{Shuicheng Yan}
	\affiliation{%
		 \institution{Sea AI Lab}
		 \country{Singapore}
        }
	\email{shuicheng.yan@gmail.com}
        \authornote{Corresponding author.}
	
	\begin{abstract}
		Garment sewing pattern represents the intrinsic rest shape of a garment, and is the core for many applications like fashion design, virtual try-on, and digital avatars. In this work, we explore the challenging problem of recovering garment sewing patterns from daily photos for augmenting these applications. To solve the problem, we first synthesize a versatile dataset, named SewFactory, which consists of around 1M images and ground-truth sewing patterns for model training and quantitative evaluation. SewFactory covers a wide range of human poses, body shapes, and sewing patterns, and possesses realistic appearances thanks to the proposed human texture synthesis network. Then, we propose a two-level Transformer network called Sewformer, which significantly improves the sewing pattern prediction performance. Extensive experiments demonstrate that the proposed framework is effective in recovering sewing patterns and well generalizes to casually-taken human photos. Code, dataset, and pre-trained models are available at: \url{https://sewformer.github.io}.
	\end{abstract}


	
	\begin{teaserfigure}
  \includegraphics[width=\textwidth]{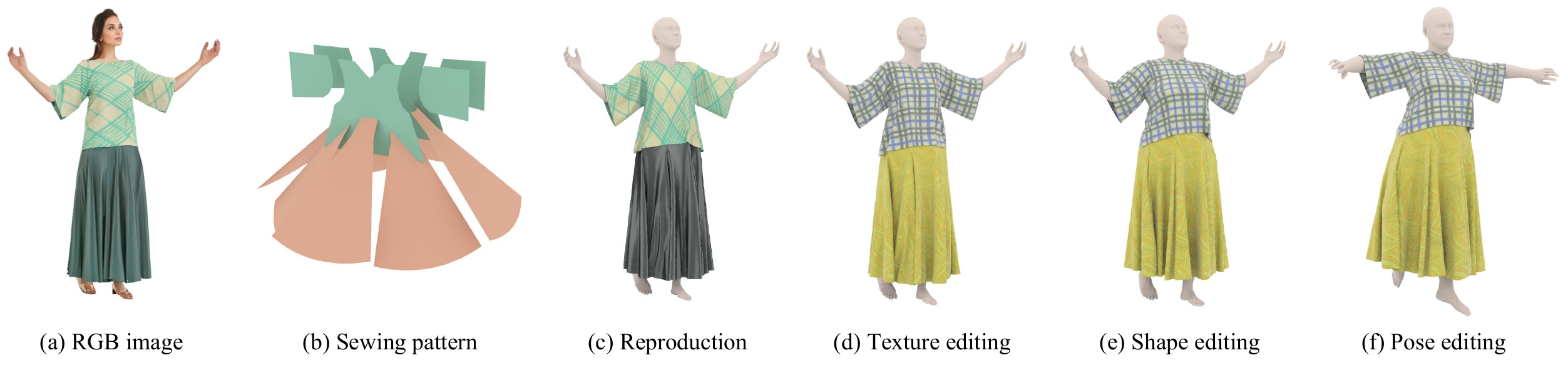}
		\caption{Given a single RGB image of a clothed human (a), the proposed algorithm can accurately recover the underlying garment sewing pattern (b), leading to wide applications in virtual/augmented reality, for example, 3D garment mesh reconstruction (c) and 3D garment editing in terms of garment texture (d), human shape (e), and human pose (f).
		}
		\Description{figure description}
		\label{fig:teaser}
	\end{teaserfigure}

	\maketitle
	
	\section{Introduction} 
	In this paper, we study the problem of estimating garment sewing patterns from a single RGB image ({Fig.~\ref{fig:teaser}}).
	A garment sewing pattern~\cite{bang2021estimating,koro2022neuraltailor,KorostelevaGarmentData} is a collection of 2D polygons called panels that can be stitched together to form a garment.
	It represents the intrinsic rest shape of a garment disentangled from the undesirable complicacy induced by extrinsic factors, such as external physical forces, collisions, and fabric properties. 
	Moreover, it is parametric and thus allows direct and interpretable control over the garment design.

	Existing works on sewing pattern reconstruction mainly rely on rich 3D information as input, such as high-quality 3D scans~\cite{bang2021estimating} or point clouds~\cite{koro2022neuraltailor} which are not accessible for general users, limiting their applications in many practical scenarios.
	In this work,  we focus on a more challenging task: estimating sewing patterns from a single image, which can be obtained with only a regular camera or from monocular human photos on the Internet.
	The estimated sewing pattern enables more accessible garment manipulation, such as clothes draping on novel body poses, and flexible editing of the shape of a captured garment, and further benefits many downstream applications in metaverse, such as virtual try-on, garment design, and avatar creation.
	However, this problem is very challenging and is still less explored due to both the inherent ill-posedness of 2D-to-3D conversion and the complicacy of real-world conditions, such as varied camera views, challenging human poses, and occlusions.

	Currently, among the few works in this direction, most of them adopt optimization-based approaches for sewing pattern reconstruction from images~\cite{jeong2015garment,yang2018physics}. 
	Though achieving good results by imposing heuristic rules and priors, these methods are typically slow in inference due to the time-consuming optimization process, less user-friendly due to the difficulty of hyper-parameter tuning for different images, and susceptible to real-world inputs where the manually-designed rules could be violated.
	Recently, \citet{chen2022structure} propose a deep neural network for recovering sewing patterns.
	Nevertheless, this method does not consider the variations in garment types, human poses, and textures of real-world photos, and thereby cannot generalize well to in-the-wild data.

	In this work, we target a data-driven framework for efficient and generalizable sewing pattern reconstruction using a single image.
	There are two major challenges for building such a framework.
	First, there is a lack of suitable training data. 
	Existing garment datasets either do not have sewing pattern annotations~\cite{zhu2020deep} or are insufficiently diverse and realistic in terms of clothes appearances and human poses~\cite{KorostelevaGarmentData}.
	To close this gap, we synthesize a new large-scale dataset, called SewFactory. SewFactory consists of one million image-and-sewing-pattern pairs with diverse garments under a wide range of human shapes and poses, facilitating more effective model training and evaluation.
	In order to synthesize such a dataset, a common challenge faced by many recent datasets~\cite{bertiche2020cloth3d,KorostelevaGarmentData} is that the rendered human body lacks photorealistic textures.
	A simple solution to this problem is to directly apply pre-scanned human textures~\cite{varol17_surreal} onto human meshes. However, this approach often leads to low-quality appearances and artifacts due to poor scan quality, low human diversity, and 3D discontinuity in UV mapping.
	To address this issue, we develop a novel neural network for textured human synthesis, which enhances our dataset by adding more realistic skin, faces, and hair.
	Besides, SewFactory has abundant high-quality annotations, including depth maps, 3D human shapes and poses, garment meshes and textures, and 3D semantic segmentation labels, which could potentially benefit other tasks in Metaverse beyond this work.

	The second challenge is that the data structures of sewing patterns are highly irregular and vary significantly across different samples. Specifically, different garments may consist of different numbers of panels, and different panels may be enclosed by different numbers of edges. Moreover, estimating the stitch information, i.e., how individual edges of panels are connected to each other, further complicates the problem.
	To handle this issue, we propose a two-level Transformer network, called Sewformer, which aligns more closely with the data structure of sewing patterns. It separately learns the panel and edge representation in a hierarchical manner and thereby achieves higher-quality results than the existing baseline models.
	In addition, we propose a new panel shape loss as well as an SMPL-based regularization loss for network training, which helps the proposed Sewformer learn from the large amount of training data  more effectively.

	To summarize, our contributions are as follows:
	\begin{itemize}
  \item We study the challenging problem of sewing pattern reconstruction from a single unconstrained human image. 
  We make early explorations to solve this problem and introduce Sewformer, a novel two-level Transformer network, which achieves high-quality results for sewing pattern reconstruction from an image. 
\item To facilitate effective model training and evaluation, we present SewFactory, a versatile and realistic dataset comprising approximately one million image-and-sewing-pattern pairs. This dataset offers diverse clothing styles and human poses, enabling improved model performance.

  \item To improve the quality of our data, we develop a novel method for human texture synthesis.
  This method generates diverse and photorealistic human images under challenging poses, contributing to the realism and diversity of our dataset.
  \item We propose new loss functions that improve the training of Sewformer, further enhancing the quality of the reconstructed garment panels.
	\end{itemize}
	
		\begin{table*}[t]
  \small
		\begin{center}
			\caption{\textbf{Comparison of the proposed SewFactory dataset with existing datasets.} SewFactory is the first large-scale dataset that provides sewing pattern annotations along with high pose variation and realistic garment and human textures. ``\#Garment'' denotes the number of garment instances. ``Pose Var'' indicates the level of pose variation exhibited in the dataset, with ``None'' denoting that all human is in a fixed T-pose. ``G-Texture Var'' and ``H-Texture Var'' represent garment texture and human texture variations, respectively.}
			\label{tab:dataset}
			\newcommand{\resize}{\text}
			\definecolor{lightblue}{rgb}{0.8, 0.92, 1}
			\begin{tabular}{lcccccc}
				\toprule 
				\resize{\textbf{Dataset}} &  \resize{\textbf{Real/Syn}}  & \resize{\textbf{\#Garment}} &  \resize{\textbf{Pose Var}} & \resize{\textbf{Sewing Pattern}} & \resize{\textbf{G-Texture Var}} & \resize{\textbf{H-Texture Var}} \\
				\midrule 
				MGN~\cite{bhatnagar2019multi} & Real  & 712 & Low & \ding{56} & Low & Low \\
				DeepFasion3D~\cite{zhu2020deep} & Real & 563 & Low & \ding{56} & Low & None \\
				3DPeople~\cite{pumarola20193dpeople} & Syn  & 80 & High & \ding{56} & Low & Low \\
				Cloth3D~\cite{bertiche2020cloth3d} & Syn & 11.3k & High & \ding{56} & High & Low \\
				Wang \text{et al.}~\cite{garmentdesign_Wang_SA18} & Syn  & 8k & None & \ding{52} & Low & None \\
				Korosteleva and Lee~\cite{KorostelevaGarmentData} & Syn  & 22.5k & None & \ding{52} & Low & None \\
				\cellcolor{lightblue}SewFactory & \cellcolor{lightblue}Syn  & \cellcolor{lightblue}19.1k & \cellcolor{lightblue}High & \cellcolor{lightblue}\ding{52}& \cellcolor{lightblue}High & \cellcolor{lightblue}High \\
				\bottomrule 
			\end{tabular}
		\end{center}

	\end{table*}
	
	\section{Related Work}
	
	\subsection{Garment Sewing Pattern Reconstruction}
	Existing works for garment sewing pattern reconstruction can be roughly categorized into 3D-based~\cite{hasler2007reverse,bang2021estimating,chen2015garment,goto2021data,koro2022neuraltailor,liu20183d} and image-based approaches~\cite{jeong2015garment,yang2018physics,chen2022structure,garmentdesign_Wang_SA18}, depending on their input type. 
	
 \paragraph{3D-based sewing pattern reconstruction.}
	Early approaches in this direction use either template matching~\cite{hasler2007reverse,chen2015garment} or surface flattening~\cite{bang2021estimating,sharp2018variational} for sewing pattern reconstruction from point clouds or 3D meshes of dressed humans.
	More recently, some researchers have explored data-driven methods for recovering sewing patterns from 3D data~\cite{goto2021data,koro2022neuraltailor}.
	\citet{goto2021data} propose a deep convolutional neural network (CNN)~\cite{isola2017image} together with exponential map~\cite{schmidt2006interactive} for garment panel reconstruction.
	\citet{koro2022neuraltailor} propose NeuralTailor, which employs a hybrid network architecture for recovering the garment panels and stitching information from point clouds.
	While these methods have achieved good results, their reliance on high-quality 3D input data limits their usage in many applications where 3D sensors are not accessible.

 \begin{figure}
		\centering
		\includegraphics[width=1.0\linewidth]{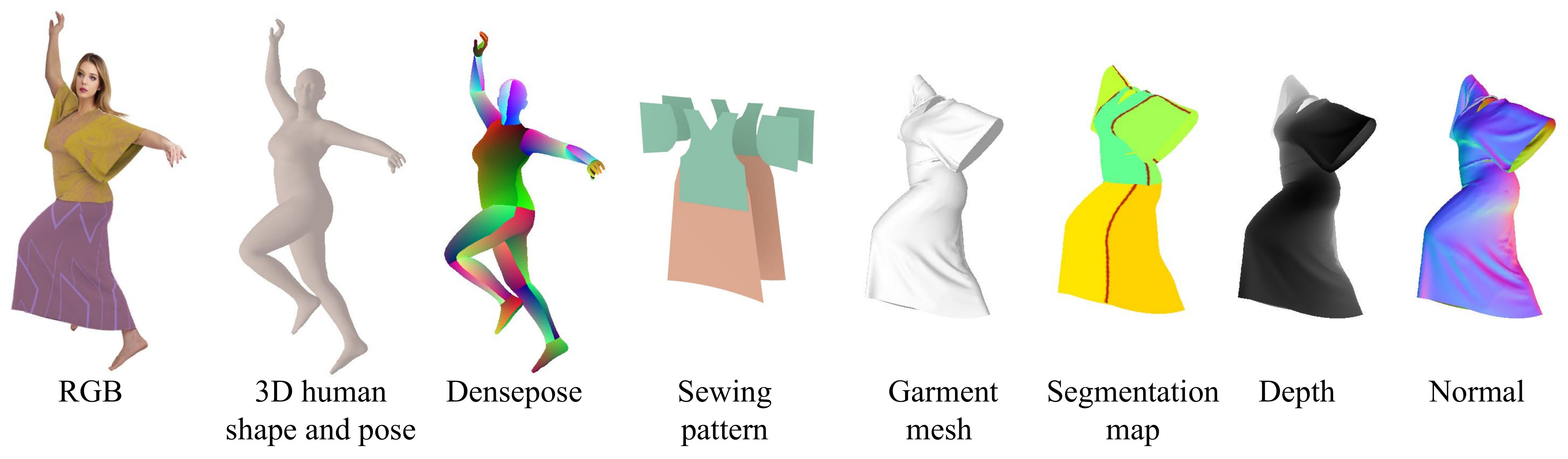}
		\caption{\textbf{Annotations of the SewFactory dataset.} 
			We generate around one million RGB images where each is annotated with diverse ground-truth labels in the figure, supporting a wide variety of tasks in computer vision and graphics.}
		\Description{figure description}
		\label{fig:dataset_all_render}
	\end{figure}

 \paragraph{Image-based sewing pattern reconstruction.}
	Due to the intrinsic difficulties of sewing pattern reconstruction from 2D data, there are only a limited number of studies in this 
	area~\cite{jeong2015garment,yang2018physics,garmentdesign_Wang_SA18,chen2022structure}.
	\citet{jeong2015garment} estimate sewing patterns from a single image by predicting the garment type and the primary body sizes via searching in a predefined database.
	However, this method is less efficient due to the exhaustive search process and does not generalize well for poses that are not in the database.
	\citet{yang2018physics} recover sewing patterns by estimating panel parameters with iterative optimization~\cite{kennedy1995particle}. 
	Nevertheless, this method requires different templates for different garment types and relies on tedious preprocessing and registration steps, which can be both computationally intensive and error-prone.
 \citet{garmentdesign_Wang_SA18} propose an encoder-decoder network for sewing pattern recovery from sketch images. 
	However, this approach requires training separate models for different garments, making it less practical for handling a wide range of garment types in real-world scenarios. 
	In addition, it is primarily designed for sketch images and cannot be easily applied to natural human photos.
	More recently, \citet{chen2022structure} propose a CNN-based model that can predict garment panels from a single image for multiple garment types.  
        To handle the irregular data structure of various garment panels, this method employs PCA to simplify the panel data structure. 
        However, this approach is only designed for predefined panel groups, restricting the output garment space.
	Moreover, due to the limitation of existing garment datasets, it only works well for human images in a standard T-pose, and its performance degenerates significantly for in-the-wild images that are captured under unconstrained poses.
	Unlike the above methods, we propose a new model called Sewformer, which together with the proposed SewFactory dataset, can effectively handle the irregular structures of various garment panels without PCA compression and generate high-quality results for  unconstrained human images with diverse shapes and poses.
	
  \paragraph{3D garment mesh reconstruction from images}
      This work is also related to recent works on image-based 3D garment mesh reconstruction~\cite{wang2018pixel2mesh,saito2019pifu,saito2020pifuhd,zhu2020deep,patel2020tailornet,tiwari2021deepdraper,zhao2021learning,moon20223d}.
	 These methods mainly train deep neural networks to directly regress the garment mesh from images, which can well reconstruct the coarse garment shape but struggle to produce realistic geometry details.
      Moreover, most of these works suffer from over-smoothing artifacts in the areas invisible to the camera.
	 In contrast, our work provides a new paradigm for image-based 3D garment mesh reconstruction, which first recovers sewing patterns from the image and then generates 3D garment mesh with the recovered patterns using an off-the-shelf simulation engine.
	 This new paradigm is closer to the real physical process of constructing a posed garment. Therefore, it can achieve higher-quality 3D garment reconstruction with realistic geometry details and produce physically plausible results even for the occluded areas. This paradigm is also highly flexible for garment editing and character animation.

	\subsection{Garment Datasets} 
	Garment data plays a key role in building data-driven models for sewing pattern recovery.
	There are mainly two types of garment datasets: 3D scanning and physical simulation.
	The first type is constructed by capturing real garments with 3D scanners~\cite{zhang2017detailed,bhatnagar2019multi,ma2020learning,zhu2020deep,tiwari2020sizer}.
	While possessing realistic appearances and dynamics, these datasets are limited in size due to the high cost of 3D scanning. 
	For example, 
 a typical dataset of this kind,
 DeepFasion3D~\cite{zhu2020deep}, comprises only 563 scanned garments, which is relatively small for training generalizable deep neural networks.
	The second type of garment datasets uses physical simulation engines for data synthesis~\cite{pumarola20193dpeople,jiang2020bcnet,bertiche2020cloth3d,hewitt2023procedural}, which overcomes the limitations of 3D-scan-based datasets and allows for the generation of larger amounts of garment data at a much lower cost. 
	Nevertheless, most of these datasets lack sewing pattern labels and thus are not suitable for our task.
	Among the few existing datasets that do provide sewing patterns, \cite{narain2012adaptive} and \cite{garmentdesign_Wang_SA18} do not have a diverse range of garment panels;
	although \cite{KorostelevaGarmentData} provides various panel templates with different topologies, it does not consider the complex textures and materials of real-world garments and only simulates the garments on a fixed T-pose human model without a realistic human texture, which leads to a significant domain gap with real-world images, hindering the generalization ability of trained models.

	To address these issues, we contribute a new synthetic garment dataset by sampling more sewing patterns and simulating the garments on a wide range of human shapes and poses. 
	We augment the dataset by adding realistic garment textures and material properties to mimic daily human photographs. 
	We also introduce a human texture synthesis network to generate diverse and realistic human appearances, which further improves the quality of our data.
	The proposed dataset provides various high-quality ground-truth labels, such as sewing patterns, segmentation masks, 3D human shape and pose, and garment meshes, making it useful for a wide variety of applications in fashion research and industry.

	\subsection{Textured Human Synthesis} 
	As mentioned above, we propose a human texture synthesis network to generate photorealistic human images for our garment dataset. 
	Since generating realistic textures plays an important role in many applications, such as human avatar creation, it has attracted great interest in recent years~\cite{xian2018texturegan,sarkar2020neural,wang2019example,zhang2020cross,lassner2017generative,weng2020misc,fu2022stylegan,sarkar2021humangan,albahar2021pose,sarkar2021style,xiang2022dressing}.
	
Some works for this problem aim to generate realistic textured human with clothed human images as exemplars based on the full-body sketch, segmentation mask, or pose-aware representations~\cite{xian2018texturegan,sarkar2020neural,zhang2020cross,albahar2021pose,sarkar2021style}. Meanwhile, another line of work synthesizes textured human from some pre-defined attributes~\cite{weng2020misc} or from randomly sampled noise with GAN architecture~\cite{fu2022stylegan,sarkar2021humangan}. 

Recently, some works focus on generating texture maps that are 3D-aware or can be directly utilized on 3D human meshes~\cite{alldieck2019learning,bhatnagar2019multi,huang2020arch,lazova2019360,saito2020pifuhd,zhao2020human,han2019clothflow,chaudhuri2021semi,chen2022auv,grigorev2021stylepeople,yang20223dhumangan,xu20213d}. Among them, reconstruction-based methods~\cite{alldieck2019learning,bhatnagar2019multi,huang2020arch,lazova2019360,saito2020pifuhd,zhao2020human} aim to predict the 3D geometry~(e.g. normal) from RGB images. Some methods instead generate 3D-aware textures~\cite{han2019clothflow} or full-body textured humans~\cite{grigorev2021stylepeople,yang20223dhumangan} by using garment templates or body shape and pose. 

	A major difference between these methods and our work is that to build the garment dataset, we are supposed to keep the simulated garments unchanged while synthesizing the human textures for the target pose.
	This introduces additional challenges, as we need to generate humans with realistic skin and hair while maintaining challenging or even out-of-distribution poses without altering their corresponding garments. To fulfill these requirements, we devise a novel human texture synthesis network that introduces a learnable texture encoding, which facilitates effective texture extraction and warping.


	\section{SewFactory Dataset} 
	\label{sewfactory}
\begin{figure}
		\centering
		\includegraphics[width=1.0\linewidth]{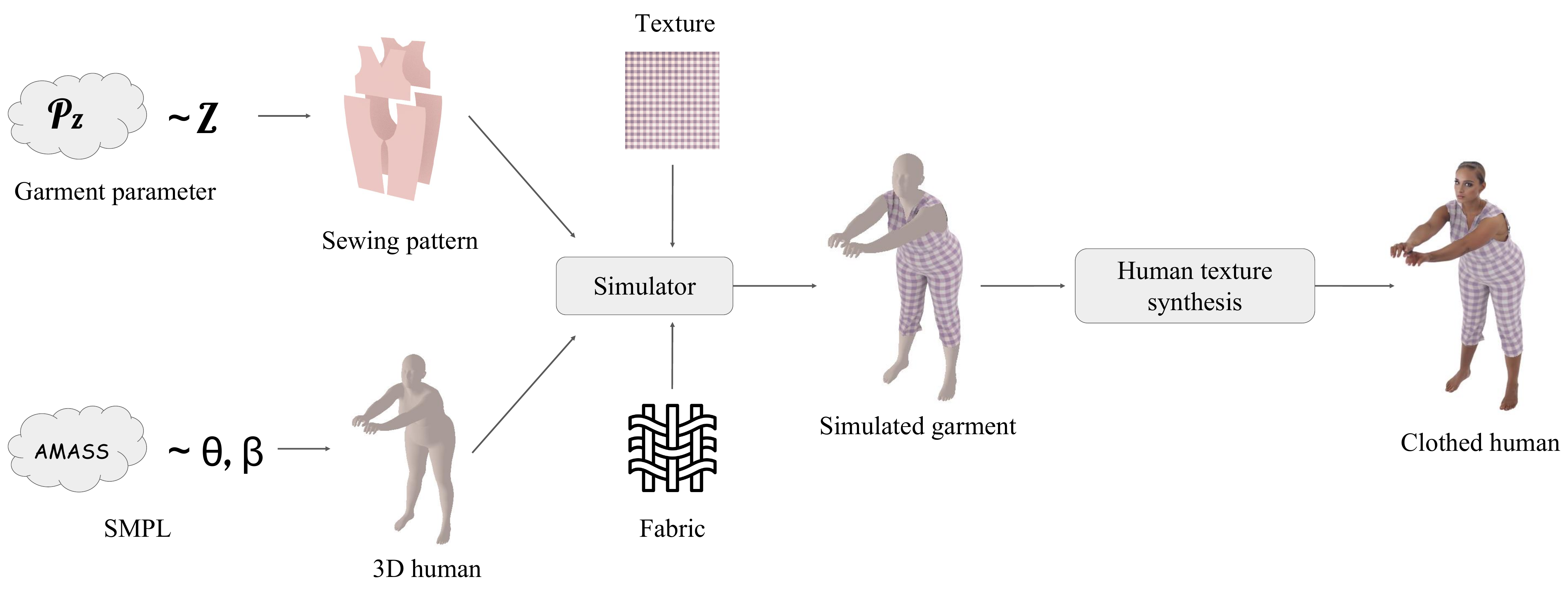}
		\caption{\textbf{Overview of the proposed data synthesis pipeline.} We first generate high-quality garments under different human shapes and poses with physical simulation and then synthesize realistic human body textures with a novel neural network.
		}
		\Description{figure description}
		\label{fig:dataset_pipline}
	\end{figure}

	We present a new dataset, SewFactory, for sewing pattern recovery from a single image.
	A comprehensive comparison between SewFactory and other existing garment datasets can be found in Table~\ref{tab:dataset}.
	Notably, SewFactory possesses high pose variability and a diverse range of garments and human textures, which effectively closes the domain gap with real-world inputs. 
	Moreover, SewFactory provides abundant ground-truth labels as shown in Fig.~\ref{fig:dataset_all_render}, which could potentially benefit many applications even beyond the task in this work. The whole pipeline for dataset generation is shown in Fig.~\ref{fig:dataset_pipline}, which consists of two main steps: garment simulation and human texture synthesis.
	
	\subsection{Garment Simulation}\label{sec: panel_def}
 In this step, we start by sampling a set of garment parameters, including sewing patterns, textures, and fabrics, as well as human body parameters. These parameters are then used to synthesize garments with a physical simulator~\cite{qualoth}.

	We use the templates in \cite{KorostelevaGarmentData} to generate the sewing patterns, which cover a wide range of garment shapes and topologies.
	We first uniformly sample template parameters such as the sleeve length and hem width, and then generate the sewing pattern of each garment with the templates similar to \cite{KorostelevaGarmentData}.
	A sewing pattern is composed of two parts: a group of $N_P$ panels $\{P_i\}_{i=1}^{N_P}$ and their stitching information $S$.
	Each panel $P_i$ is a closed 2D polygon encompassed by a list of $N_{i}$ edges $\{E_{i,j}\}_{j=1}^{N_{i}}$. 
	Each edge $E$ is a Bezier curve that can be represented by four scalars $x,y,c_x,c_y$, where $(x,y)$ is the 2D vector from the start to the end of the edge, and $(c_x,c_y)$ is the control point of the Bezier curve defined in a local coordinate system.
	In addition, each panel $P_i$ is associated with a 3D rotation $R_i \in \text{SO}(3)$ and a translation $T_i \in \mathbb{R}^3$, which describe the relative pose and position between the panel $P_i$ and the human body. $R_i$ and $T_i$ are used in the physical simulation process of cloth draping.
	The stitching information $S$ is defined by pairs of edges $\{(i,j),(i',j')\}$ indicating that the $i$-th panel's $j$-th edge is sewn with the $i'$-th panel's $j'$-th edge.

To simulate a diverse range of garment data, we assign a unique set of material parameters to each sample, corresponding to a unique fabric object and textile image.
We utilize three commonly-used fabric presets, namely cotton, silk, and velvet, as the basic fabric types.
Each fabric encompasses various properties specific to that fabric type, and we introduce random perturbations to the property values to simulate garments with different levels of shininess, elasticity, and other fabric-specific characteristics.
In order to enhance the appearance diversity of the simulated garments, we curate a collection of approximately 600  textile images with free license.
These images serve as the source texture, to which we apply random augmentations, such as scaling and color jittering, to expand the range of visual appearances.

	We use the Qualoth simulator~\cite{qualoth} to simulate the garment dynamics. 
	During the simulation, we randomly pair up the top and bottom garments, such as T-shirts and pants, to form garment sets, and then drape each garment set onto a unique 3D human model.
	The 3D human model is parameterized by SMPL~\cite{loper2015smpl}, which uses two parameters $\beta \in \mathbb{R}^{10}$ and $\theta \in \mathbb{R}^{24\times 3}$ to control the 3D human shape and pose.
	To ensure a sufficient amount of variation in pose, we sample 13.7k poses from the AMASS dataset \cite{mahmood2019amass} which are further interpolated into more poses using Maya~\cite{maya}.
 For each pose, we randomly sample a shape parameter $\beta$ from a uniform distribution within the range of [-1.5, 1.5].
	This is in sharp contrast to existing  datasets~\cite{garmentdesign_Wang_SA18,KorostelevaGarmentData} that only consider garments under a fixed T-pose template, leading to a significant domain gap with real-world photos.

        For each simulated garment, we render 24 views from cameras uniformly distributed around the human body using Arnold renderer~\cite{georgiev2018arnold}.
 In some rare cases, the simulator produces inappropriate simulation results, such as with wrong sizes or self-intersection, which are manually removed.
	Overall, the garment simulation leads to around one million RGB images with paired sewing pattern labels, where 85k images are used for testing, and the remaining is used for training. 
	Our data splitting ensures no garment or pose is repeated in training and testing.

	\subsection{Human Texture Synthesis} 
 \label{sec:human_texture}
\begin{figure}
		\centering
		\includegraphics[width=1.0\linewidth]{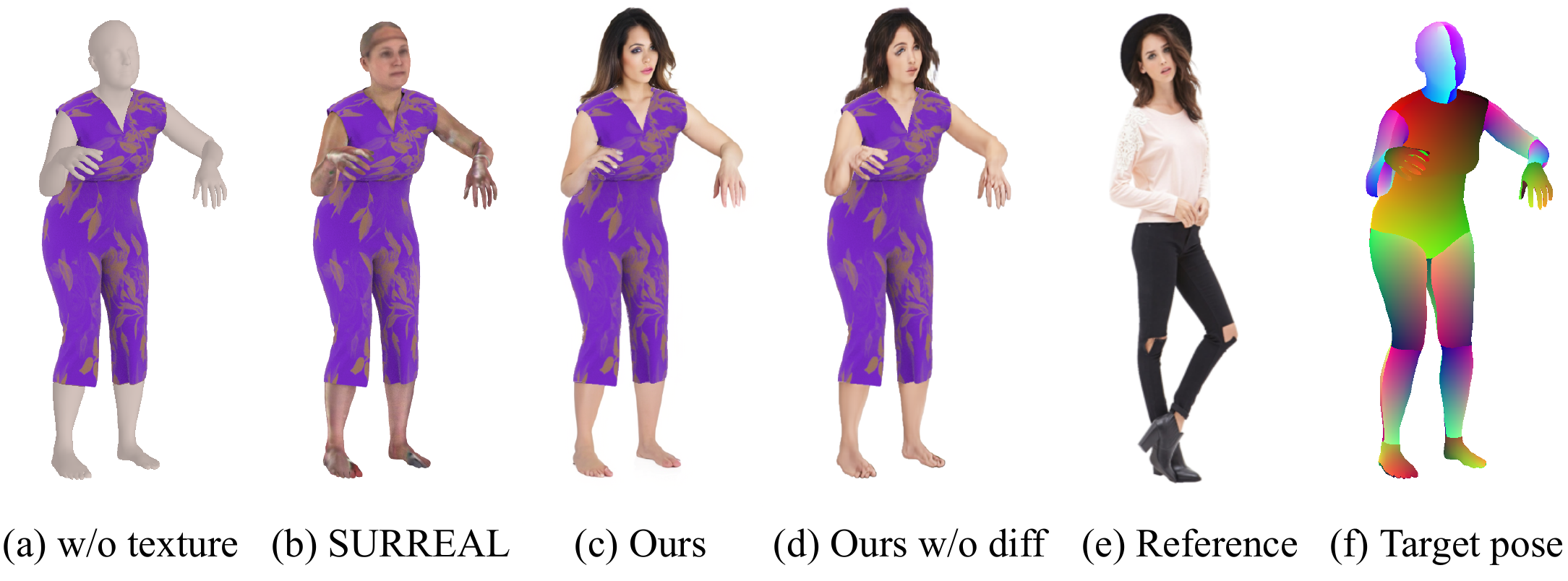}
		\caption{\textbf{Comparison of different human textures.} 
			(a) Original simulation output that does not have human texture; (b) textured human by using pre-scanned SURREAL texture~\cite{varol17_surreal}; (c) our result; (d) our result without diffusion editing. 
			Instead of creating human textures from scratch, we synthesize textured human by transferring the texture of a reference image (e) to the target pose (f).
		}
		\Description{figure description}
		\label{fig:texture_synthesis}
	\end{figure}

	While the proposed simulation system is able to produce high-quality garments and sewing patterns, the generated images still suffer from an important shortcoming that the rendered human body lacks photorealistic textures (Fig.~\ref{fig:texture_synthesis} (a)). 
	Note that this is a common challenge faced by many recent datasets~\cite{bertiche2020cloth3d,KorostelevaGarmentData}.
	A simple solution to this problem is to directly apply pre-scanned human textures, e.g., SURREAL~\cite{varol17_surreal}, onto human meshes.
	However, this approach often leads to low-quality results (Fig.~\ref{fig:texture_synthesis} (b)) due to noise and artifacts in scanning, limited human diversity, and 3D discontinuities in UV mapping.
	To address this issue, we develop a deep neural network for human texture synthesis, which enhances our dataset by adding more realistic skin, faces, and hair. 
	As shown in Fig.~\ref{fig:texture_synthesis} (c), this network allows us to greatly improve the realism and overall quality of the generated images.
	Instead of creating human textures from scratch, we utilize images of real humans~\cite{liu2016deepfashion} as a reference. This simplifies our task by reducing it to transferring the texture of a real human (Fig.~\ref{fig:texture_synthesis} (e)) to the target pose (Fig.~\ref{fig:texture_synthesis} (f)).
	Specifically, given a reference image $R_\text{img} \in \mathbb{R}^{3\times H \times W}$ where $H$ and $W$ are the height and width of the image, we aim to extract the appearance information of $R_\text{img}$ and then apply it to the target pose $T_\text{pose} \in \mathbb{R}^{3\times H \times W}$. 
	We use Densepose~\cite{guler2018densepose} to represent the target pose, as we find it better represents the semantic information of different pixels of the human body.

	One straightforward approach to this texture transfer task is to directly concatenate $R_\text{img}$ and $T_\text{pose}$ and then feed the concatenated input into a deep CNN to generate the desired human image. 
	However, this method has a major drawback in that it cannot accurately preserve the target pose, resulting in artifacts due to mismatch with the simulated garment.
	Moreover, it tends to overfit training poses and degenerates severely on out-of-distribution poses in test data.
	To address this issue, we propose to first extract the texture map of the human body and then warp the texture to the target pose.
	An overview of the proposed framework for textured human synthesis is shown in Fig.~\ref{fig:sim2real_framework}, which is comprised of three stages as below.

	\subsubsection{Neural texture extraction.}
\begin{figure*}
		\centering
		\includegraphics[width=1.0\textwidth]{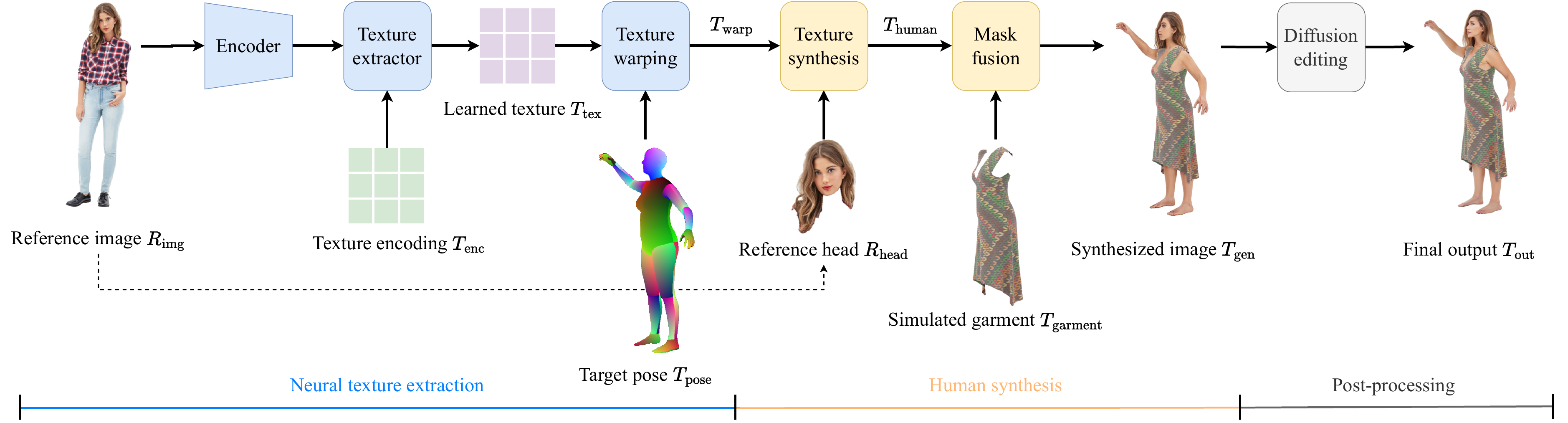}
		\caption{\textbf{Overview of the proposed framework for human texture synthesis.} First, we use a shared texture encoding $T_\text{enc}$ to extract texture features $T_\text{tex}$ from the reference image $R_\text{img}$ and then warp these features to the target pose $T_\text{pose}$. Next, we generate textured human $T_\text{gen}$ by first mapping the warped features $T_\text{warp}$ to RGB human textures $T_\text{human}$ and then fusing $T_\text{human}$ with the simulated garment $T_\text{garment}$ with mask fusion.
		Finally, we employ a diffusion editing module to further refine the synthesized result.}
		\Description{figure description}
		\label{fig:sim2real_framework}
	\end{figure*}

	At the core of the texture extraction stage is the texture extractor in Fig.~\ref{fig:sim2real_framework}.
	It  maps a texture encoding $T_\text{enc} \in \mathbb{R}^{D \times N_\text{p} \times H_\text{tex} \times W_\text{tex}}$ to a texture map $T_\text{tex} \in \mathbb{R}^{D \times N_\text{p} \times H_\text{tex} \times W_\text{tex}}$ by aggregating the information from the reference image $R_\text{img}$, where $H_\text{tex},W_\text{tex},D$ are the height, width, and number of channels of the texture map, and $N_\text{p}$ is the number of body parts defined in Densepose~\cite{guler2018densepose}.
	$T_\text{tex}$ can be seen as a generalization of the UV map in \cite{xu20213d}, which describes the 3D textures of different body parts. 
	$T_\text{enc}$ is a learnable encoding tensor that is randomly initialized and shared across different samples.
	Similar to \cite{xu20213d}, we first use a deep CNN encoder to convert $R_\text{img}$ into feature space. Then texture extractor exploits the cross attention mechanism to non-locally distribute the information of $R_\text{img}$ into $T_\text{tex}$.
	Next, we warp the learned texture features $T_\text{tex}$ to the target pose $T_\text{pose}$ with bilinear sampling, which generates the warped features $T_\text{warp} \in \mathbb{R}^{D\times H \times W}$.
	Our neural texture extraction approach ensures the synthesized human body always conforms to the target pose, resulting in a proper fit for the simulated garment.
	In addition, as the texture extractor is decoupled from the target pose, our method is less susceptible to out-of-distribution test poses.
 
	\subsubsection{Human synthesis.}
	While one can use $T_\text{warp}$ as the final output (by setting $D=3$), it often leads to low-quality head regions as the Densepose is not capable of precisely depicting the human head.
	Instead, we generate the textured human $T_\text{human}$ with another deep CNN (``Texture synthesis'' in Fig.~\ref{fig:sim2real_framework}) under the guidance of the head part  of the reference image, denoted by $R_\text{head}$.
	We adopt the U-Net architecture~\cite{ronneberger2015u} for the texture synthesis CNN, and the guidance information of $R_\text{head}$ is injected with the adaptive normalization of StyleGAN~\cite{karras2019style}.
	With the synthesized human body $T_\text{human}$, we can generate the clothed human image with mask fusion:
	\begin{align}
		T_\text{gen} = T_\text{human} \cdot (1-\text{Mask}) + T_{\text{garment}} \cdot \text{Mask}, \nonumber
	\end{align}
	where $T_\text{garment}$ and \text{Mask} are the garment texture and mask obtained from the simulation system. 
	
	We train our network with a combination of pixel-wise loss, perceptual loss~\cite{johnson2016perceptual}, feature-matching loss~\cite{xu2017learning}, and adversarial loss~\cite{goodfellow2020generative}. 
	
	\subsubsection{Post-processing.}
\begin{figure}
		\centering
		\includegraphics[width=1.0\linewidth]{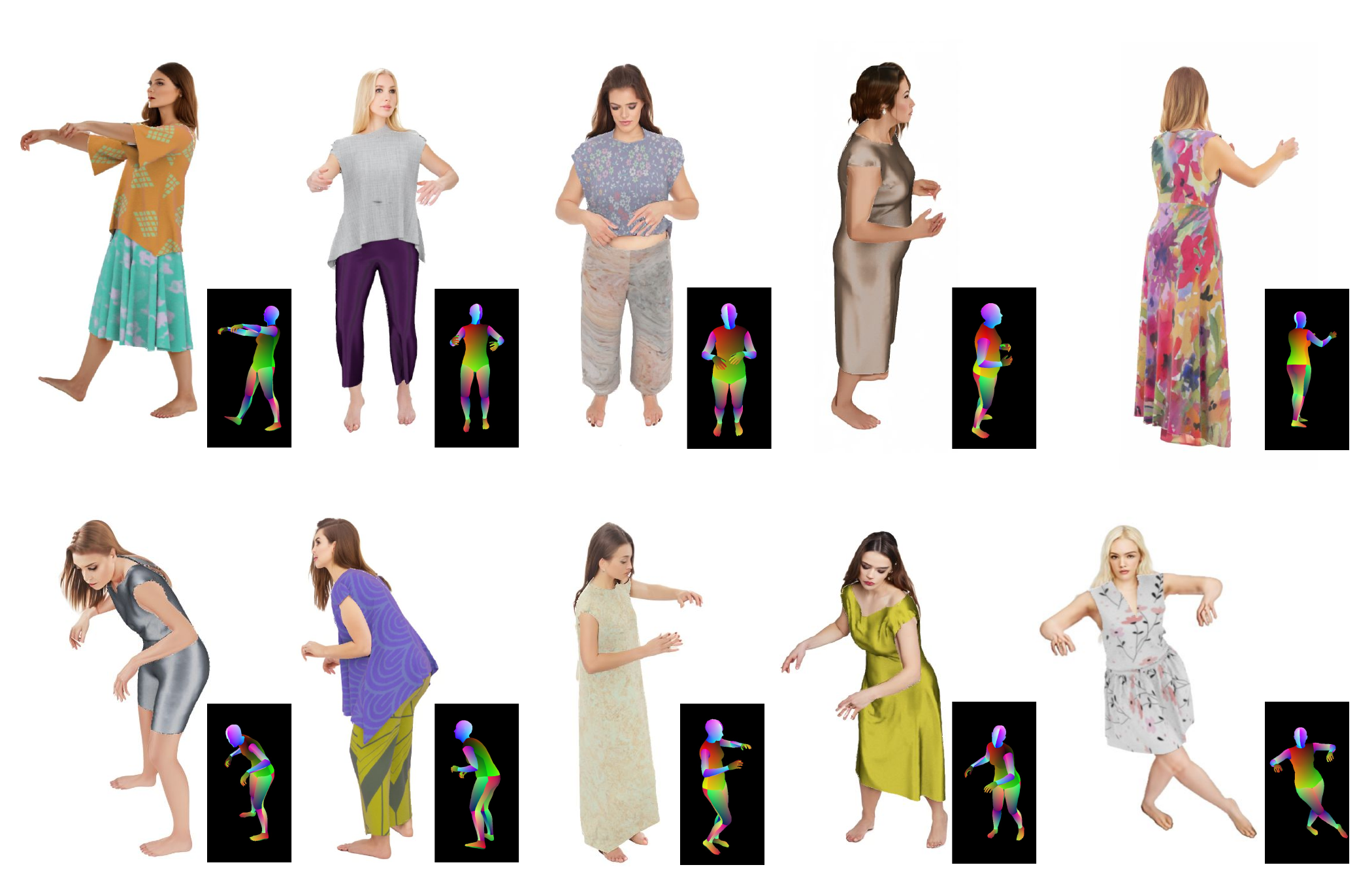}
		\caption{\textbf{Example results of human texture synthesis.} The proposed algorithm generalizes well to various challenging poses in the black boxes.}
		\Description{figure description}
		\label{fig:sim2real_examples_main}
	\end{figure}

	As shown in Fig.~\ref{fig:texture_synthesis} (d), the direct output $T_\text{gen}$ from the human synthesis network still contains a considerable amount of artifacts and distortions. 
	To further improve the quality of the generated human, we employ diffusion models~\cite{rombach2022high} as a powerful image prior for post-processing. 
	Specifically, we follow the SDEdit framework~\cite{meng2021sdedit} to refine the human images. We first perturb the synthesized image $T_\text{gen}$ with a moderate ratio of Gaussian noise and then progressively remove the noise with a denoising network, which effectively improves the realism of the final output $T_\text{out}$ (Fig.~\ref{fig:texture_synthesis} (c)) and closes the gap with real human photos.
	Eventually, the proposed data generation pipeline enables highly diversified results with various poses and realistic appearances as shown in Fig.~\ref{fig:sim2real_examples_main}.
	
	\section{Sewing Pattern Reconstruction}
	
	\begin{figure*}
		\includegraphics[width=0.98\textwidth]{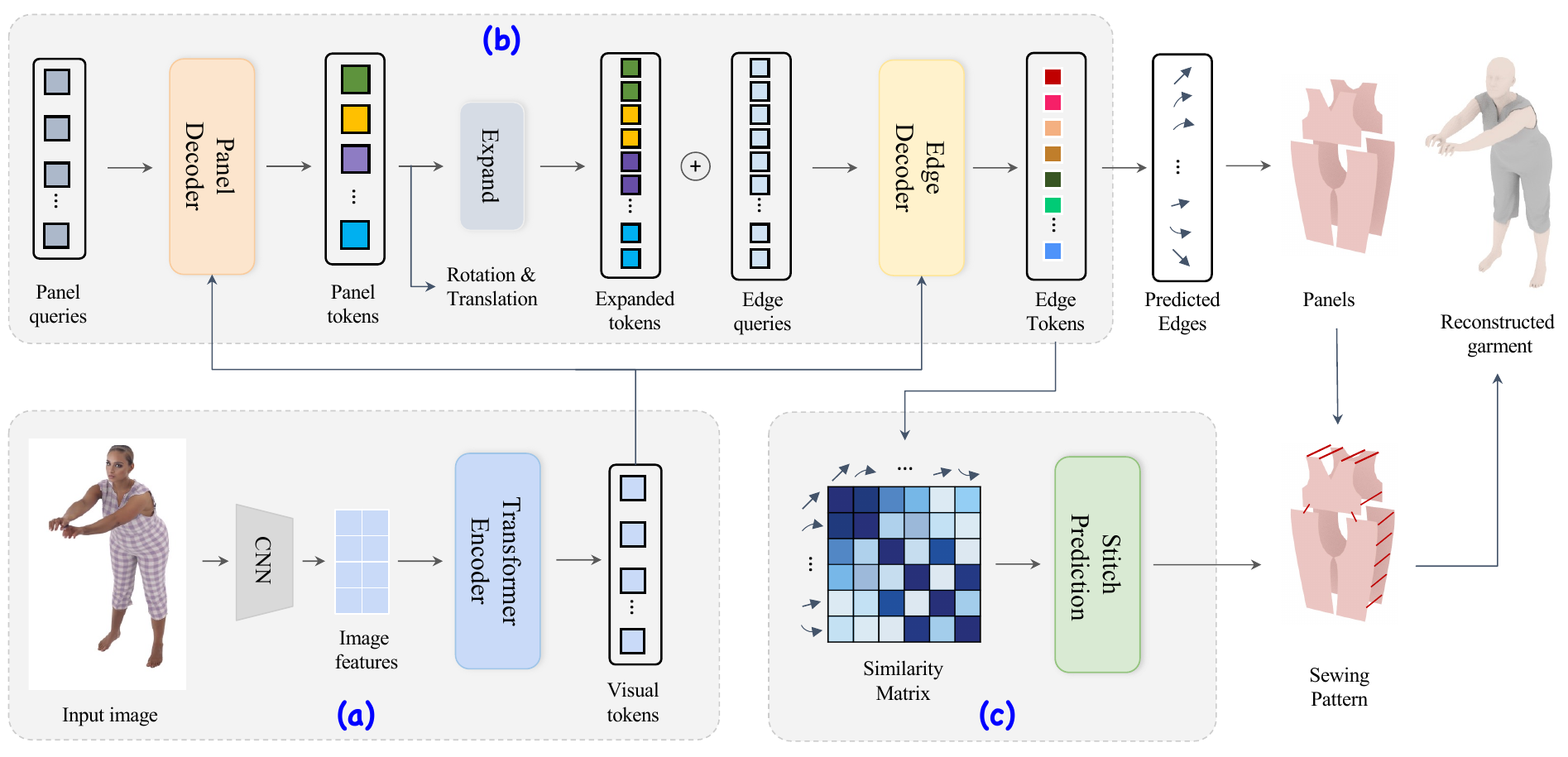}
		\caption{\textbf{Overview of the proposed Sewformer}. We first send the input image into a visual encoder (a) to generate a sequence of visual tokens. 
			Then the visual tokens are fed into a two-level Transformer decoder (b) to produce panel-level and edge-level feature tokens, which are subsequently used to recover the garment panels.
                The edge features are also used to estimate the stitching relations via the stitch prediction module (c).
		}
		\Description{figure description}
		\label{fig:network}
	\end{figure*}

	As introduced in Section~\ref{sec: panel_def}, the sewing pattern in our SewFactory dataset has a highly irregular data structure, making it unsuitable for the commonly-used deep CNNs that are primarily designed for regular data structures. 
        Instead, we propose a new model, called Sewformer, to better accommodate this data. We also propose new loss functions to facilitate training of Sewformer.
	
	\subsection{Architecture} \label{sec:sewformer}
	An overview of Sewformer is shown in Fig.~\ref{fig:network}.
	It consists of three main components: (a) a visual encoder to learn sequential visual representations from the input image, (b) a two-level Transformer decoder to obtain the sewing pattern in a hierarchical manner, and (c) a stitch prediction module that recovers how different panels are stitched together to form a garment. 

        One of the key designs of our work is the two-level structure of the Transformer decoder. It provides a simple yet effective way to handle garment sewing patterns. This design is novel and has not been previously applied to the task of garment reconstruction. 
        
	\subsubsection{Visual encoder.}
	To use Transformers for sewing pattern reconstruction, we employ the visual encoder to convert the input image into sequential data. 
	Specifically, we first generate a low-resolution feature map $F \in \mathbb{R}^{C \times H_F \times W_F}$ from the input image with a CNN backbone (ResNet-50~\cite{he2016deep} in our experiments). 
	Then we serialize $F$ by reshaping it into ${C \times H_F  W_F}$ where the spatial dimensions are flattened in 1D.
	These serialized features are subsequently processed by a Transformer encoder to learn the visual tokens  $F_\text{vis} \in \mathbb{R}^{C \times H_F  W_F}$, where each encoder layer consists of a standard multi-head self-attention module and an MLP similar to \cite{vaswani2017attention}. 

	\subsubsection{Two-level Transformer decoder}
	Based on the learned visual tokens $F_\text{vis}$, 
    we propose a two-level Transformer decoder to recover the garment panels $\{P_i\}_{i=1}^{N_P}$ that are introduced in Section~\ref{sec: panel_def}.
	As shown in Fig.~\ref{fig:network}, the first level (panel decoder) is designed to extract the overall information of the panels; the second level (edge decoder) is dedicated to learning the specific shapes of the panels by recovering the edge information.
	As will be shown in Section~\ref{sec:experiment}, compared to normal Transformers that only have a one-level decoder, the proposed Sewformer considers the two-level (panel and edge) characteristics of sewing patterns and is able to more effectively recover the garments in a coarse-to-fine manner.
	
	For the panel decoder, we first randomly initialize $N_P$ panel queries $\{Q_\text{P}^i\}_{i=1}^{N_P}$ with Gaussian distribution. These panel queries are trained using backpropagation along with the whole model.
        We predict the panel tokens $\{F_P^i\}_{i=1}^{{N_P}}$ by applying a multi-head cross-attention layer between the queries $Q_P^i$ and the visual tokens $F_\text{vis}$:
	\begin{align} \label{eq:cross_attention}
            {F_P^i}=\text{CrossAttention}(Q_\text{P}^i, F_\text{vis}).
	\end{align}
	Then we can use an MLP to predict the 3D rotation and translation of the panels, i.e., $R_i, T_i = \text{MLP}({F_P^i})$.
	
	Similar to the panel decoder, we initialize a set of random query embeddings for all the panel edges in the edge decoder which are learned in training.
    We use a maximum number of panel and edge queries and remove invalid predictions during inference (edges with lengths near zero and panels with fewer than three edges). 
	Then we combine each edge query with its corresponding panel feature using element-wise sum and pass the combined embeddings into an MLP to obtain the final edge queries.
	In this way, the edge queries belonging to the same panel share the same panel feature, which facilitates producing consistent edges.
	
        Furthermore, the coarse-to-fine learning process, which progresses from panel to edge, effectively alleviates the difficulties in training, while directly learning a large number of edges could be overwhelming and lead to undesirable local minimums.
        
	Similar to Eq.~\ref{eq:cross_attention}, we feed the edge queries and visual tokens into a cross-attention layer to generate each edge token $F_{E}^{ij}$ (corresponding to the $j$-th edge of the $i$-th panel).
	The edge tokens are then used in an MLP to predict the Bezier edge parameter $E_{i,j} = \text{MLP}({F_E^{ij}})$.

	\subsubsection{Stitch prediction.}
        Our stitch prediction module is similar to that of NeuralTailor~\cite{koro2022neuraltailor}.
	As introduced in Section~\ref{sec: panel_def}, the stitches are defined by pairs of edges from different panels. 
	  Since stitched two edges typically have similar features, such as orientations and spatial locations, we predict whether two edges form a stitch based on the similarity between the edge features. 
	As shown in Fig.~\ref{fig:network}, we first construct the similarity matrix between all pairs of edges in the predicted sewing pattern using the learned edge tokens as stitch tags and then obtain the stitch predictions by iteratively finding the maximum values in the matrix. 
    Specifically, since the similarity matrix is symmetric, we first remove its lower triangular part. We identify the maximum value of the remaining entries and then eliminate the corresponding row and column. This process is iterated until all the stitches are determined.

	\subsection{Loss Function}
	
	Our training objective is composed of three parts: a panel prediction loss,  a stitch prediction loss, and an SMPL-based regularization term:
	
	\begin{equation}
		\label{eq:total_loss}
		\mathcal{L}_{\text{total}} = \lambda_1 \mathcal{L}_{\text{panel}} + \lambda_2 \mathcal{L}_{\text{stitch}} + \lambda_3 \mathcal{L}_{\text{SMPL}},
	\end{equation}
        where $\lambda_1, \lambda_2, \lambda_3$ are hyperparameters to balance each term. We use a stitch prediction loss $\mathcal{L}_{\text{stitch}}$ similar to \cite{koro2022neuraltailor}, and the other two terms are explained below.
	
	\subsubsection{Panel prediction loss} \label{sec:shape_loss} The panel prediction loss consists of three terms: 1) the shape loss $\mathcal{L}_{\text{shape}}$ to encourage high-fidelity shapes of the reconstructed panels; 2) the loop loss $\mathcal{L}_{\text{loop}}$ to enforce that the edges of a panel form a closed loop; 3) the rotation and translation loss $\mathcal{L}_{\text{RT}}$ to encourage accurate 3D rotation and translation predictions:
	\begin{equation}
		\label{eq:panel_loss}
		\mathcal{L}_{\text{panel}} = \mathcal{L}_{\text{shape}} + \mathcal{L}_{\text{loop}} + \mathcal{L}_{\text{RT}}, 
	\end{equation}
        where $\mathcal{L}_{\text{loop}}$ and $\mathcal{L}_{\text{RT}}$ are defined the same way as NeuralTailor~\cite{koro2022neuraltailor}.
 
\begin{figure}
		\includegraphics[width=\linewidth]{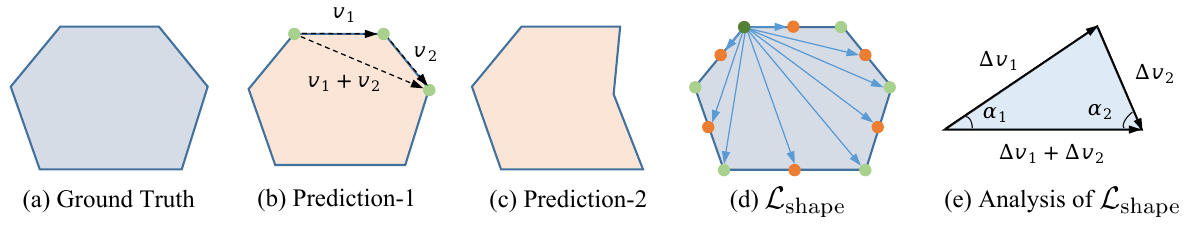}
		\caption{\textbf{Illustration of the proposed panel shape loss.} (a) depicts a ground truth panel. (b) Each edge of Prediction-1 is slightly shorter than the corresponding edge in the ground truth. (c) While most edges of Prediction-2 match the ground truth exactly, the two right-side edges exhibit significant errors. Overall, (b) and (c) have the same per-edge loss $\mathcal{L}_\text{shape-NT}$~\cite{koro2022neuraltailor}. (d) shows the support vectors connecting vertices on the edges, including the endpoints (green) and midpoints (orange). We only show the support vectors emanating from the dark green vertex and omit others for clarity.
        (e) provides an intuitive analysis of $\mathcal{L}_\text{shape}$, showing how it encourages more evenly distributed errors among edges of the panel.}
		\Description{figure description}
		\label{fig:shapeloss}
	\end{figure}

	For the shape loss, a straightforward choice is to use the one in NeuralTailor~\cite{koro2022neuraltailor} as well,
        which directly penalizes the L2 distance between the predicted edge and the ground-truth edge (solid blue lines in Fig.~\ref{fig:shapeloss}(a)). 
    Nevertheless, we find that this is not always an adequate measure of shape discrepancy between different panels.
    For instance, despite the shape of Prediction-1 in Fig.~\ref{fig:shapeloss}(b) being closer to the ground truth in Fig.~\ref{fig:shapeloss}(a) than Prediction-2 in Fig.~\ref{fig:shapeloss}(c), the two predictions yield the same value for the per-edge loss $\mathcal{L}_{\text{shape-NT}}$. 

    An important cause for this problem is that the per-edge loss only provides 1D comparisons (lines) between the prediction and the ground truth, resulting in sparse and implicit supervision of the 2D shapes. To explicitly enforce shape similarity in 2D, a better solution is to convert the panel edges into binary 2D masks and penalize the discrepancy between these masks. However, this conversion involves rasterization operations that are non-differentiable, resulting in difficulties in training.

To address this issue, we propose a novel shape loss $\mathcal{L}_{\text{shape}}$ to approximate the 2D mask loss, as illustrated in Fig.~\ref{fig:shapeloss}(d). 
We start by collecting a set of vertices on the panel edges and sampling support vectors on the panel by connecting each pair of vertices. Our shape loss $\mathcal{L}_{\text{shape}}$ is then defined as the L2 error between the support vectors in the predicted panels and those in the ground-truth panels.
$\mathcal{L}_{\text{shape}}$ can be seen as a densified version of the per-edge loss $\mathcal{L}_{\text{shape-NT}}$, which better encourages 2D shape similarity with the ground truth. 
As expected, the more support vectors that are sampled, the better our shape loss approximates the 2D mask loss. In our implementation, we use the endpoints and midpoints of the panel edges as the vertices for simplicity.

To more intuitively understand why sampling more vectors on the panel leads to a better shape loss, we provide a toy example of two edge vectors on the panel, \textit{i.e.}, $v_1$ and $v_2$ in Fig.~\ref{fig:shapeloss}(b).
Denoting the ground-truth vectors as $\hat{v}_1$ and $\hat{v}_2$, the per-edge loss could be written as:
\begin{align} \label{eq:toy_shape_nt}
    \|v_1 - \hat{v}_1\| + \|v_2 - \hat{v}_2\| =  \| \Delta v_1 \| + \| \Delta v_2 \|,
\end{align}
where we define $\Delta v=v-\hat{v}$, and $\|\cdot\|$ represents the L2 norm.
Correspondingly, the proposed $\mathcal{L}_{\text{shape}}$ becomes:
\begin{align} \label{eq:toy_shape}
    &\|v_1 - \hat{v}_1\| + \|v_2 - \hat{v}_2\| + \|(v_1 + v_2) - (\hat{v}_1+\hat{v}_2)\| \nonumber \\
    =& \| \Delta v_1 \| + \| \Delta v_2 \| + \| \Delta v_1 + \Delta v_2 \| \nonumber \\
    =& (1+\cos \alpha_1) \| \Delta v_1 \| + (1+\cos \alpha_2) \| \Delta v_2 \|,
\end{align}
where $v_1+v_2$ is the newly-added support vector in Fig.~\ref{fig:shapeloss}(b), and the angles $\alpha_1$ and $\alpha_2$ are defined in Fig.~\ref{fig:shapeloss}(e).
By comparing Eq.~\ref{eq:toy_shape_nt} and \ref{eq:toy_shape}, we can see that the proportion between the two errors $\| \Delta v_1 \|$ and $\| \Delta v_2 \|$ are adjusted by the proposed shape loss.
Suppose that the prediction of $v_1$ has a larger error than $v_2$, \textit{i.e.}, $\| \Delta v_1 \| > \| \Delta v_2 \|$, then we have $(1+\cos \alpha_1) / (1+\cos \alpha_2) > 1$, implying that larger errors are more heavily penalized in Eq.~\ref{eq:toy_shape}. 
In other words, $\mathcal{L}_{\text{shape}}$ encourages more evenly distributed errors among all edges, and thereby the prediction in Fig.~\ref{fig:shapeloss}(b) is preferred over Fig.~\ref{fig:shapeloss}(c). 

For simplicity of the explanation, we assume the predicted edges are lines instead of Bezier curves in Eq.~\ref{eq:toy_shape_nt}. However, the underlying concept remains valid, as curves can be effectively approximated by piece-wise lines.%

	\subsubsection{SMPL-based regularization loss.} \label{sec:smpl_loss}

 To further improve the estimation of 3D garment panels, a good understanding of the 3D human pose in the input image could be beneficial.
 As the garment panels are typically closely related to the shape and movement of the body, the 3D human pose that describes the position and orientation of body parts can provide vital information to infer the shape and location of the garment panels in 3D space, even when they are partially visible or occluded in the input image.

Motivated by this idea, we introduce an SMPL-based regularization loss term in Eq.~\ref{eq:total_loss} to guide the training process of Sewformer.
Specifically, we add an extra set of pose queries to the panel queries in Fig.~\ref{fig:network}(b), which leads to an additional output of 3D human pose $\theta$ after the panel decoder. 
Then the regularization term $\mathcal{L}_\text{SMPL}$ is defined as the mean squared error between the predicted 3D human pose and the ground truth.
As the learned pose features are adaptively blended into panel tokens by the attention mechanism in the panel decoder, $\mathcal{L}_\text{SMPL}$ essentially facilitates garment panel reconstruction with human pose information.
Note that we did not supervise the human shape $\beta$ as we empirically find no benefits in our experiments.
 Thanks to the abundant labels provided by the SewFactory dataset, we can easily apply the SMPL regularization term in a supervised manner.

	\begin{figure*}
		\includegraphics[width=.95\textwidth]{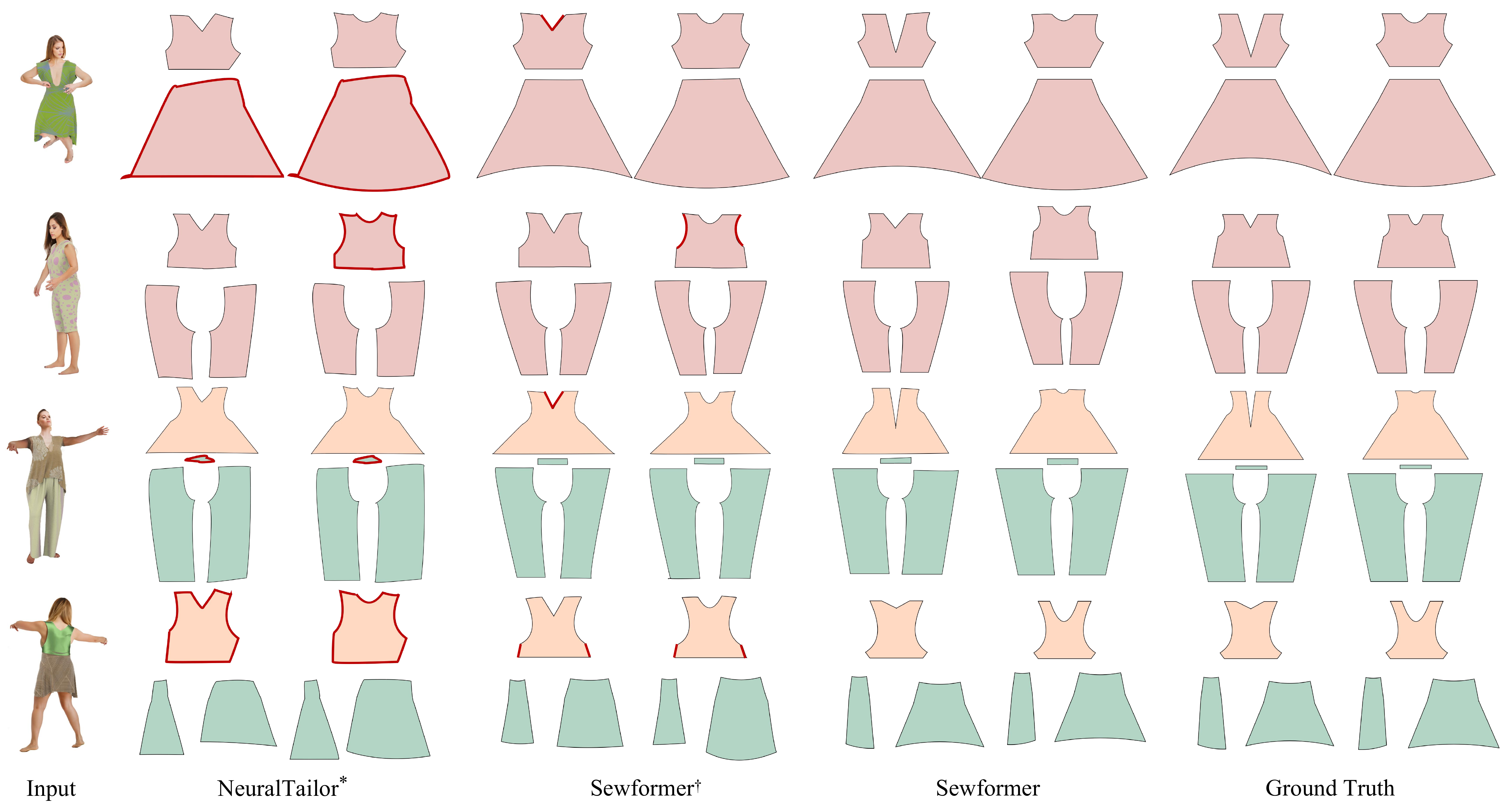}
		\caption{\textbf{Qualitative evaluation of panel predictions by different methods on our dataset.} For each method, the two columns represent the front view and back view results, respectively. Major errors of the baseline approaches are highlighted with red edges. 
  In the last row, the display order for the 4-panel skirt is left-front-right-back. Please see the caption of Table~\ref{tab:table1} for the explanations of the baseline methods.}
		\Description{figure description}
		\label{fig:comparison}
	\end{figure*}

\subsection{Relationship with NeuralTailor}

The proposed Sewformer is related to NeuralTailor~\cite{koro2022neuraltailor}, as both designs share similar elements such as the loop and RT losses in Eq.~\ref{eq:panel_loss}. 
Nevertheless, Sewformer introduces several fundamental improvements and contributions that distinguish it from NeuralTailor.
First, Sewformer utilizes a new two-level Transformer model, which effectively handles the complex data structure of sewing patterns. 
This design is a significant departure from the hybrid architecture of EdgeCNN, attention-MLP, and LSTM used in NeuralTailor.
While being much simpler, it results in better accuracy in high-quality sewing pattern reconstruction.
Second, we introduce two novel loss functions, $\mathcal{L}_\text{shape}$ and $\mathcal{L}_\text{SMPL}$, which significantly improve the training process.
In addition, different from prior methods~\cite{koro2022neuraltailor,chen2022structure} that only handle one garment at a time (upper or lower clothes), the proposed Sewformer is capable of predicting the panels for the entire clothing set in a single model run.
To separate the recovered panels into upper and lower garments, we leverage the predicted stitching relations and group connected panels into the same clothes piece.
Lastly, empowered by the SewFactory dataset, our proposed algorithm is capable of effectively handling casual human photos captured in unconstrained poses, while NeuralTailor is limited to T-pose inputs, restricting its applicability in real-world scenarios.

 \begin{table*}[h!]
		\begin{center}
			\caption{\textbf{Quantitative evaluation on the SewFactory dataset.} The evaluation metrics are introduced in Section~\ref{sec:sota}. Sewformer$^\dagger$ is a variant of our model with a one-level Transformer decoder. NeuralTailor$^\ast$ is the NeuralTailor model~\cite{koro2022neuraltailor} trained with our proposed loss functions.
   $\uparrow$: the higher the better; $\downarrow$: the lower the better. 
   }
			\label{tab:table1}
			\begin{tabular}{lccccc|ccc}
				\toprule 
				Model & Panel L2 $\downarrow$ & Rot L2 $\downarrow$ & Trans L2 $\downarrow$ & \#Panel $\uparrow$ & \#Edges $\uparrow$ &  Precision $\uparrow$ & Recall $\uparrow$ & F1 score $\uparrow$\\
				\midrule 
				Sewformer & \textbf{3.57} & \textbf{0.0205} & \textbf{0.693} & \textbf{88.7\%} & 97.5\% &   \textbf{96.1\%} & 95.4\% & \textbf{95.7\%}\\
				Sewformer$^\dagger$ & 3.91 & 0.0322 & 0.979  & $87.5\%$ & $95.6\%$ &  $82.8\%$  & $98.9\%$ & $90.1\%$\\
				NeuralTailor$^\ast$  & 4.15 & 0.0347 & 0.995 & $83.8\%$ & $97.5\%$ &  $76.8\%$& \textbf{99.6\%} &$86.7\%$\\
				NeuralTailor  &  4.41 & 0.0300 &  1.050& $83.6\%$ & \textbf{97.8\%} &  $81.5\%$& $87.8\%$ &$84.5\%$\\
				\bottomrule 
			\end{tabular}
		\end{center}
	\end{table*}
	\section{Experiments} \label{sec:experiment}

 We first describe the implementation details of the proposed Sewformer. Then we provide qualitative and quantitative evaluations of our algorithm on both synthetic and real-world images. 

\subsection{Implementation Details}
\paragraph{Dataset.}
We use the SewFactory dataset to train and evaluate our proposed Sewformer. 
We group panels of different garment types based on their spatial attributes, such as top front, top back, sleeve left front, etc., resulting in 24 semantic panel classes. 
For efficient training, we align all garment parameters and ensure that they have the same dimension by applying zero padding.

\paragraph{Training.}
During training, we apply random rotation, affine, and perspective augmentations to the input images, followed by resizing to a fixed size of $384\times 384$. We use the AdamW optimizer. We set the initial learning rate of the Transformer to $10^{-4}$, the initial learning rate of the CNN backbone to $10^{-5}$, and the weight decay to $10^{-4}$. The backbone is initialized with pre-trained weights from ImageNet, while the rest of the model is initialized randomly. We train the models for 40 epochs using 8 A100 GPUs with a total batch size of 512. To balance the different loss terms in the objective function (Eq.~\ref{eq:total_loss}), we set the hyperparameters as $\lambda_1=10$, $\lambda_2=0.5$, and $\lambda_3=1$.

\subsection{Comparison with the State of the Art}\label{sec:sota}
        We compare the proposed algorithm against NeuralTailor~\cite{koro2022neuraltailor}, the state-of-the-art method for recovering garment sewing patterns. 
        Since NeuralTailor is designed for 3D point clouds, we adapt it to our task by replacing the original graph-based encoder with a ResNet-50 architecture for feature extraction. The extracted image features are spatially flattened and treated as point features in the subsequent modules of NeuralTailor.
        As NeuralTailor is originally trained on the dataset of \cite{KorostelevaGarmentData}, it is only suitable for fixed T-pose garments and cannot handle diverse human poses captured in everyday photos.
        For a more comprehensive comparison, we retrain NeuralTailor on our proposed SewFactory dataset.
        %

        \paragraph{Evaluation metrics.}
        We use the same evaluation metrics as \cite{koro2022neuraltailor}: 
        1) Panel L2: the L2 distance between the edge parameters of the predicted and ground-truth panels, measuring the quality of shape predictions; 
        2) Rot L2 and Trans L2: the L2 error of the predicted rotations $R$ and translations $T$ compared to their ground truth; 
        3) \#Panel: the accuracy of the predicted number of panels within each garment pattern; 
        4) \#Edges: the accuracy of the number of edges within each correctly-predicted panel; 
        5) the precision, recall, and F1 score of the stitches, evaluating the quality of recovered stitching relations.

        \paragraph{Quantitative evaluation.}
        As shown in Table~\ref{tab:table1}, our Sewformer demonstrates superior performance compared to NeuralTailor across multiple evaluation metrics. 
        In particular, our method achieves a  relative decrease of 19\% in Panel L2 error, a relative decrease of 32\% in Rot L2 error, a relative decrease of 34\% in Trans L2 error, an absolute increase of 5.1\% in \#Panel accuracy, and an absolute increase of 11.2\% in F1 score, showing the effectiveness of our algorithm. 
        Meanwhile, our method exhibits a slightly lower accuracy in \#Edges compared to NeuralTailor. 
        This discrepancy arises due to the fact that our approach well recovers more panels than NeuralTailor, including panels that consist of particularly challenging edges. Consequently, these challenging edges contribute to the increased complexity of edge estimation in our \#Edges results.

        \begin{table}[t]
\footnotesize
    \centering
    \caption{\textbf{Effectiveness of the proposed loss functions.} ``w/o $\mathcal{L}_\text{shape}$'' represents the model trained with $\mathcal{L}_\text{shape-NT}$.}
    \label{table:ablation}
    \begin{tabular}{lccccc}
        \toprule
        Model & Panel L2 $\downarrow$ & Rot L2 $\downarrow$ & Trans L2 $\downarrow$ & \#Panel $\uparrow$ & \#Edges $\uparrow$  \\
        \midrule
        full model & \textbf{3.57} & \textbf{0.0205} & \textbf{0.693} & \textbf{88.7\%} & 97.5\%   \\
        w/o $\mathcal{L}_\text{shape}$ & 3.71 & 0.0260 & 0.966 & 88.2\% & 97.8\%  \\
        w/o $\mathcal{L}_\text{SMPL}$ & 3.63 & 0.0243 & 0.897 & 86.1\% & \textbf{97.9\%} \\
        \bottomrule
    \end{tabular}
\end{table}

        Furthermore, it is worth mentioning that Sewformer and NeuralTailor are trained using different loss functions. To emphasize the efficacy of the architecture of Sewformer, we also retrain NeuralTailor using our proposed loss functions (denoted as NeuralTailor$^*$ in Table~\ref{tab:table1}). Although NeuralTailor$^*$ demonstrates improvements across most metrics compared to the baseline, it still clearly falls  short of the performance of Sewformer.

\paragraph{Qualitative evaluation.}
We present qualitative evaluations in Fig.~\ref{fig:comparison}. 
It is evident that our proposed model outperforms NeuralTailor by a substantial margin. Particularly, the results obtained from NeuralTailor exhibit significant deficiencies in terms of fidelity and accuracy. For instance, the skirt appears distorted, and the waistband takes on an irregular polygonal form, deviating from the ground-truth rectangular shape.
In contrast, Sewformer produces more precise details, such as the length of the skirts and pants. 

\begin{figure}[t]
		\includegraphics[width=0.5\textwidth]{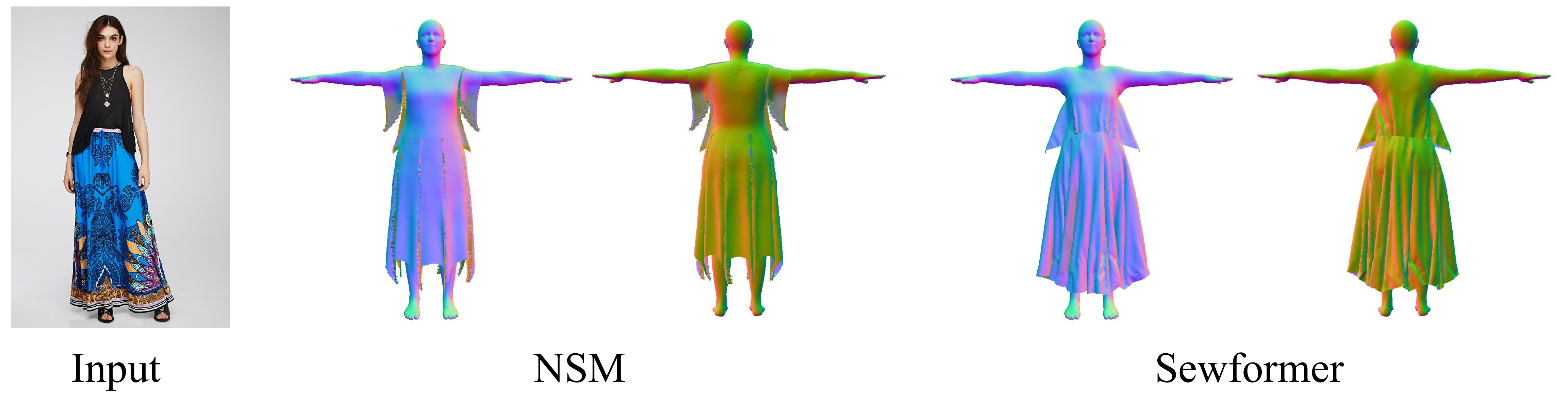}
		\caption{\textbf{Comparison with NSM}~\cite{chen2022structure}. The proposed algorithm achieves higher-quality garment reconstruction. 
  }
		\Description{figure description}
		\label{fig:compare_nsm}
	\end{figure}

\paragraph{Comparing with NSM.} 
We further compare the proposed Sewformer with NSM~\cite{chen2022structure}. 
Since the code for NSM is not publicly available, we directly use the reconstructed garment mesh provided by the original authors for comparison. 
As shown in Fig.~\ref{fig:compare_nsm}, our method produces high-quality results, while the results of NSM suffer from low fidelity and artifacts.
Moreover, these artifacts make it problematic to simulate the garment of NSM for different poses, while our results can be conveniently manipulated and edited as elaborated in Section~\ref{sec:editing}.

\subsection{Ablation Study}

\paragraph{Effectiveness of the two-level architecture.}
In Section~\ref{sec:sewformer}, we propose a two-level Transformer decoder to handle the hierarchical structure of sewing patterns.
To investigate the impact of this design, we train a variant of the Sewformer with a one-level decoder (denoted as Sewformer$^\dagger$), which relies on the panel tokens to query the garment information from the visual encoder and predicts the final edge vectors with a subsequent MLP.
As shown in Table~\ref{tab:table1}, while Sewformer$^\dagger$ gives better results than NeuralTailor, there is still a clear performance gap between Sewformer$^\dagger$ and Sewformer, highlighting the effectiveness of the two-level design.
Further, the visual examples in Fig.~\ref{fig:comparison} also support this observation,
where the results of Sewformer$^\dagger$ lack important details, especially in panel corners. In contrast, the proposed model consistently generates more accurate results, demonstrating the crucial role of the two-level Transformer in capturing intricate panel details.

\begin{figure}
		\includegraphics[width=0.95\linewidth]{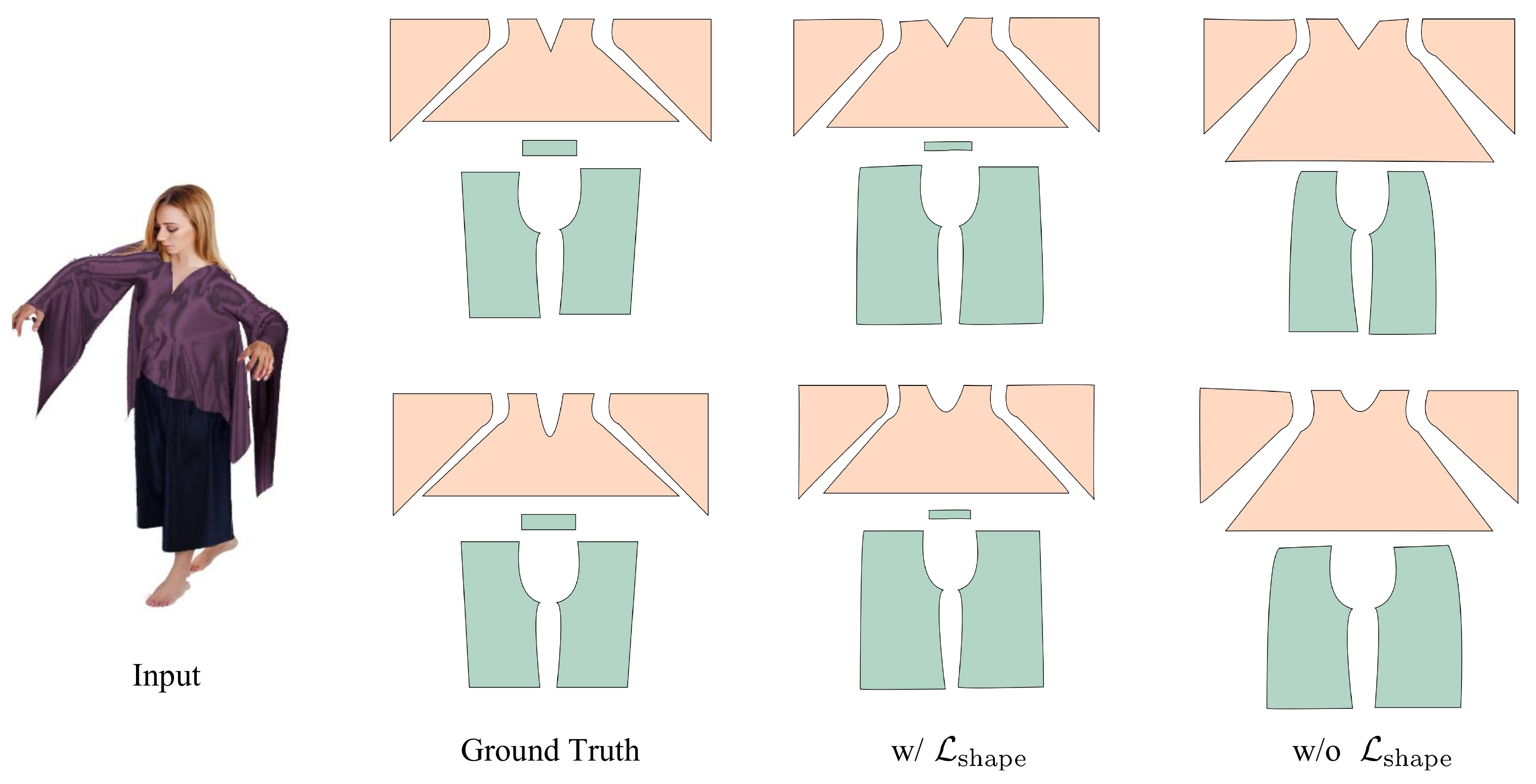}
		\caption{\textbf{Effect of different shape losses.} Compared to the prior method~\cite{koro2022neuraltailor} (w/o $\mathcal{L}_\text{shape}$), the proposed shape loss (w/ $\mathcal{L}_\text{shape}$) leads to better shape predictions closer to the ground truth. The top and bottom rows show the front and back views, respectively.}
		\Description{figure description}
		\label{fig:shape-comp}
	\end{figure}

\paragraph{Effectiveness of the loss functions.}
In Section~\ref{sec:shape_loss}, we introduce a new panel shape loss $\mathcal{L}_\text{shape}$ which better encourages shape similarity between the prediction and ground truth compared to the prior method $\mathcal{L}_\text{shape-NT}$ from \cite{koro2022neuraltailor}.
As shown in Table~\ref{table:ablation}, the proposed shape loss achieves noticeably smaller errors in panel shape (Panel L2). 
This improvement is further supported by the visual example in Fig.~\ref{fig:shape-comp}, which demonstrates that our proposed shape loss is able to recover panel shapes closer to the ground truth. 
Remarkably, $\mathcal{L}_\text{shape}$ also improves the prediction accuracy of the panel rotation and translation in Table~\ref{table:ablation}. 
We hypothesize that this can be attributed to that our shape loss encourages the model to learn more discriminative panel features, which consequently facilitates more accurate estimations of panel rotation and translation.

\begin{figure}
		\includegraphics[width=0.95\linewidth]{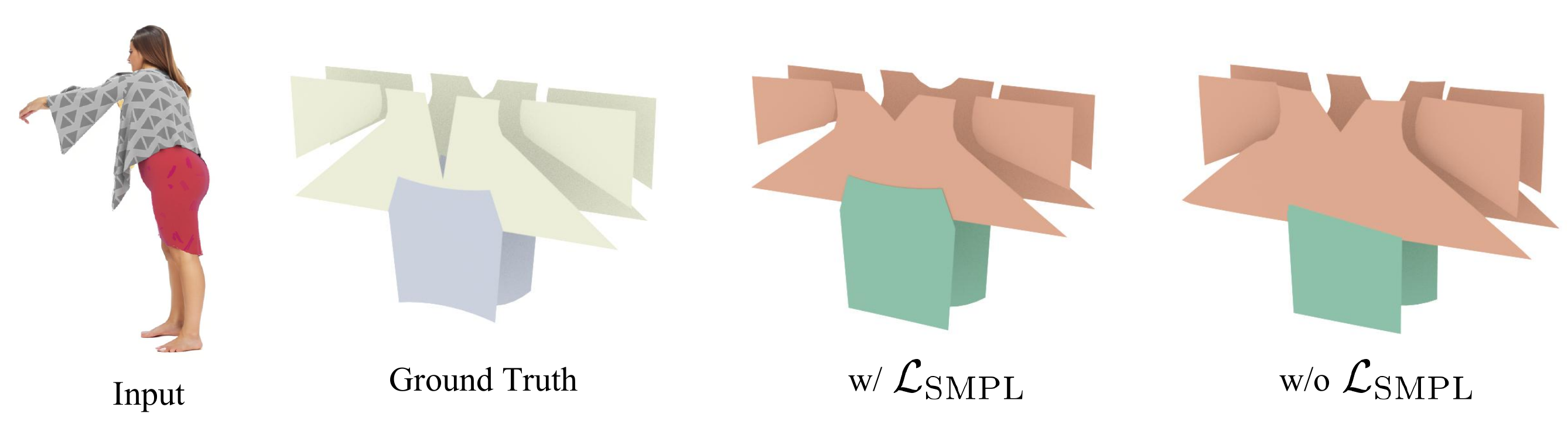}
		\caption{\textbf{Effect of the SMPL Loss.} The result without $\mathcal{L}_\text{SMPL}$ suffers from incorrect spatial arrangement of the panels and noticeable shape errors. In comparison, incorporating $\mathcal{L}_\text{SMPL}$ leads to improved performance, generating plausible results even for occluded regions.}
		\Description{figure description}
		\label{fig:smpl-comp}
	\end{figure}

As a garment is closely related to the human body wearing it, 
we propose an SMPL-based regularization loss $\mathcal{L}_\text{SMPL}$ to exploit the human body information for improving garment panel reconstruction.
As shown in Table~\ref{table:ablation}, 
the absence of this loss leads to inaccurate panel rotation and translation predictions, as well as a significant drop in the \#Panel accuracy, showing the important role of human body information in understanding the spatial relationship of different panels.
Additionally, Fig.~\ref{fig:smpl-comp} demonstrates that the model trained without the SMPL loss produces inferior results in terms of panel shape and global 3D parameters. Notably, our proposed $\mathcal{L}_\text{SMPL}$ produces plausible results even for occluded regions, emphasizing the significance of incorporating human body information for garment panel reconstruction.
%


\begin{figure*}
\includegraphics[width=.95\textwidth]{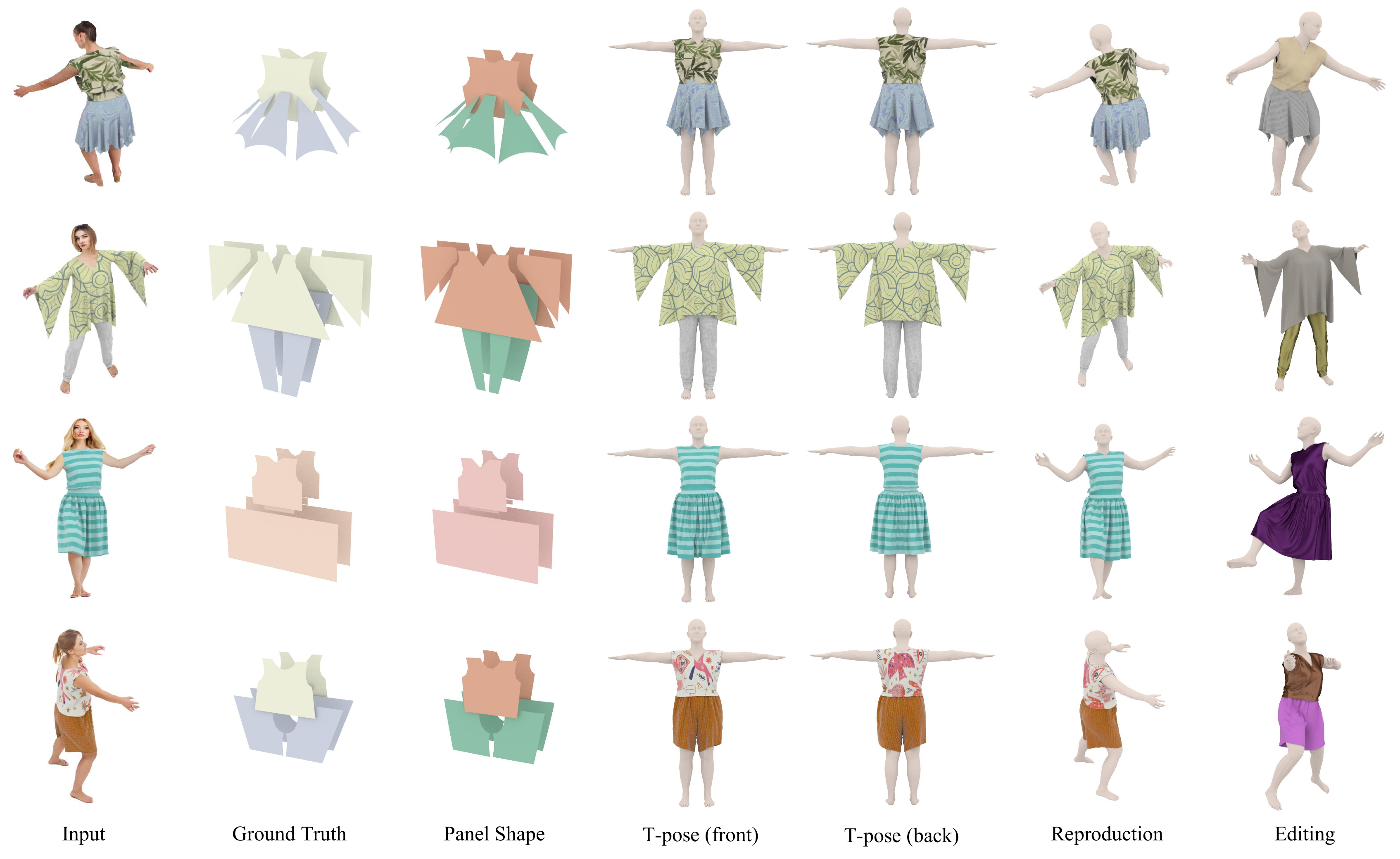}
\caption{\textbf{Garment reproduction and editing.} Each row shows an example. The first column is the input RGB image, and the second column is the corresponding ground truth sewing pattern. The third column is the sewing pattern recovered by our method. For ease of comparison, we show the back-side panels in the front for the first and fourth examples, where the human is facing backward. The fourth and fifth columns show the recovered garment in T-pose. The sixth column is the reproduction of the input garment, and the last column is a random editing result. The garment textures are manually added.}
\Description{figure description}
\label{fig:main-results}
\end{figure*}

\begin{figure}
\includegraphics[width=.99\linewidth]{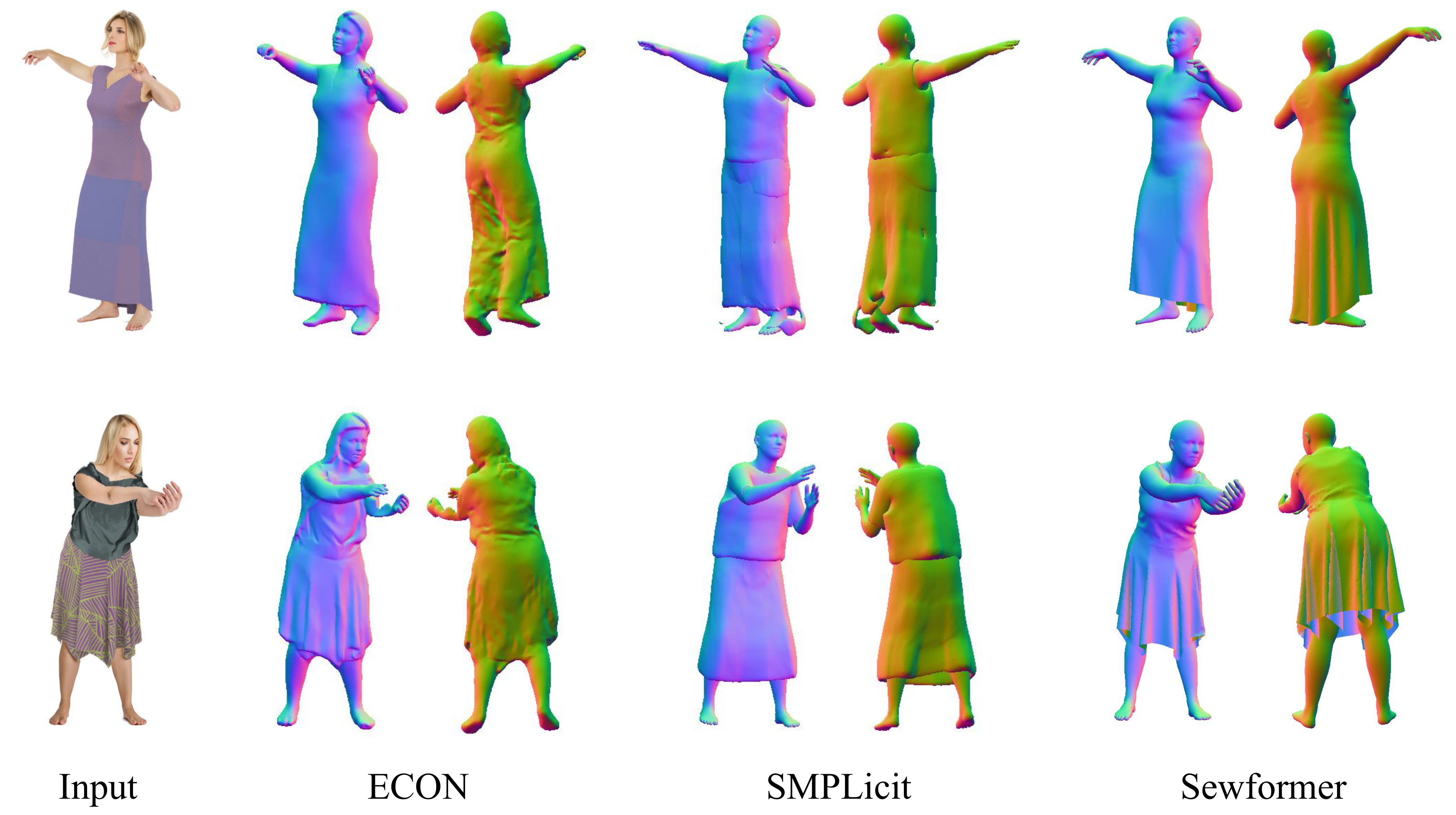}
\caption{\textbf{Comparison with single-view garment mesh reconstruction methods.} Compared to ECON~\cite{xiu2023econ} and SMPLicit~\cite{corona2021smplicit}, the proposed method achieves high fidelity and realistic details even in occluded areas.}
\Description{figure description}
\label{fig:econ_smplicit}
\end{figure}

\subsection{Garment Reproduction and Editing}\label{sec:editing}
An important application of our work is the reproduction and editing of garments from a single image.
Specifically, given an input image, we first use the proposed Sewformer to recover its sewing pattern.
As shown in Fig.~\ref{fig:main-results}, our predicted panel shapes exhibit faithful details, such as the depth and angle of the neckline, the width of the hems and sleeves, and the symmetry within and between the panels in the sewing pattern.
The high-quality sewing pattern reconstruction allows us to use physical simulators like Qualoth~\cite{qualoth} to obtain an accurate reproduction of the 3D garment by simulating the sewing pattern on the input human shape and pose. 
We estimate the 3D human shape and pose with RSC-Net~\cite{xu20203d}. 
This process enables flexible editing of the 3D garment as illustrated in Fig.~\ref{fig:main-results}, including modifications to garment textures, human poses, human shapes, and more. These capabilities are highly valuable in procedural production processes.

\begin{figure*}
		\includegraphics[width=0.9\textwidth]{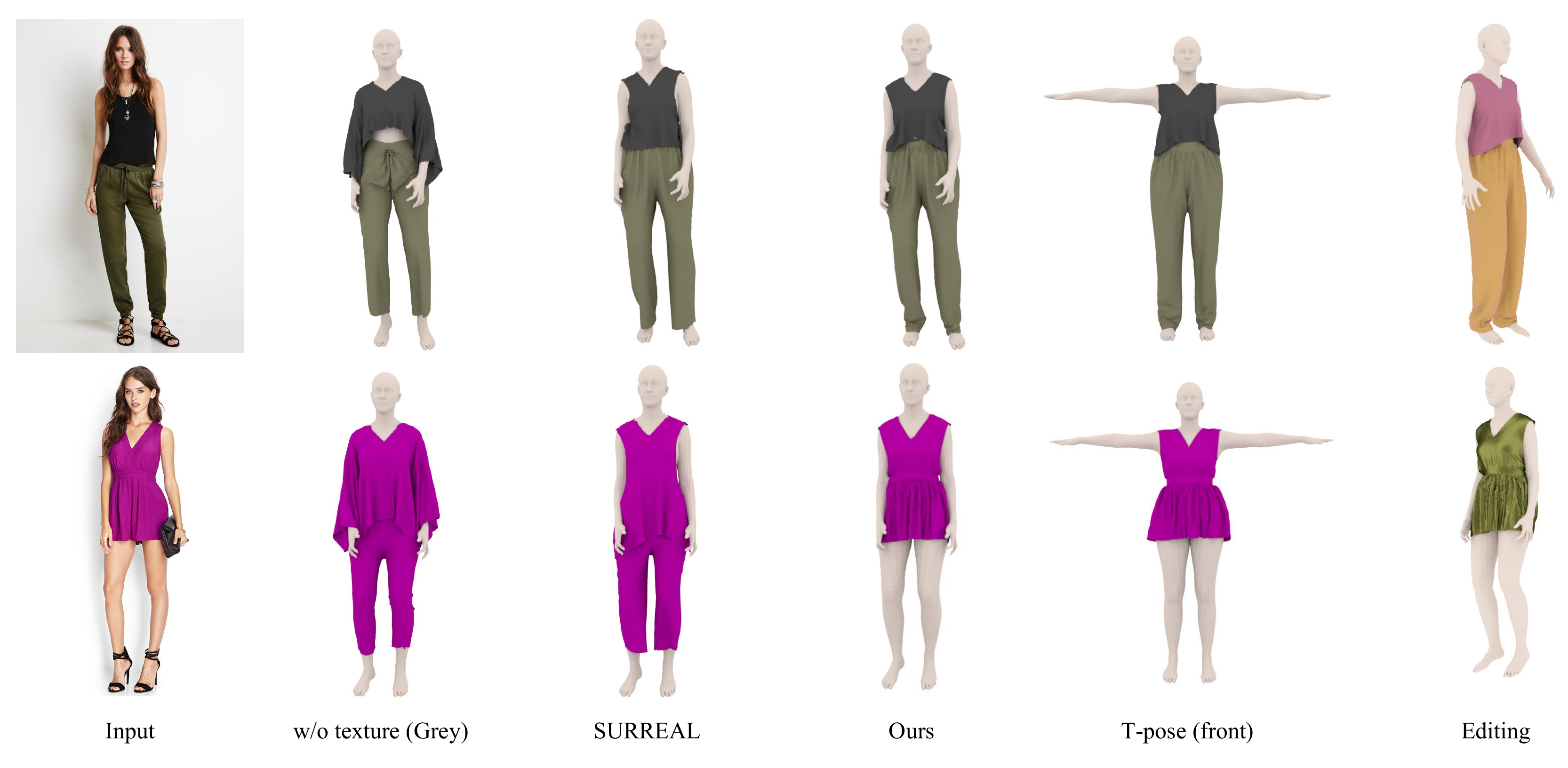}
		\caption{\textbf{Results on real human images.} 
  The second, third, and fourth columns are results recovered by the proposed network trained on three different datasets. The proposed algorithm shows strong generalization capabilities, reconstructing garments that closely resemble the input image. Meanwhile, the proposed human texture synthesis network is crucial for reducing domain gap and achieving good generalization ability.}
		\Description{figure description}
		\label{fig:deepfashon}
	\end{figure*}

\subsection{Single-View Garment Mesh Reconstruction}
In Fig.~\ref{fig:econ_smplicit}, we present a comparison with two state-of-the-art methods, ECON~\cite{xiu2023econ} and SMPLicit~\cite{corona2021smplicit}, for garment mesh reconstruction from a single image.
As shown in Fig.~\ref{fig:econ_smplicit}, while ECON well handles front views, it is susceptible to occlusions and produces unrealistic details and over-smoothed areas for the back views.
SMPLicit addresses this issue by explicitly considering clothing parameters, resulting in plausible outputs for both front and back views. However, due to the oversimplified parameter space, its results are less accurate.
In contrast, our proposed method achieves improved results, demonstrating high fidelity and realistic details even in occluded regions.
        
\subsection{Generalization to Real Photos}
We present visual results on real human images in Fig.~\ref{fig:deepfashon}. The proposed algorithm produces garments that closely resemble the input image, demonstrating the strong generalizability of our method. 

\paragraph{Effectiveness of human texture synthesis.}
In Section~\ref{sec:human_texture}, we propose a human texture synthesis network to generate realistic training data for our algorithm.
To investigate the effectiveness of the synthesized human texture, we compare the pattern reconstruction results of models trained with different human textures on real input images.
Specifically, we compare three different types of data: 1) training images without human textures; 2) training images rendered with scanned skin textures from the SURREAL dataset~\cite{varol17_surreal}; and 3) images rendered with our textures from the human texture synthesis network.
%
As ``w/o texture'' and SURREAL suffer from domain gaps with real data (Fig.~\ref{fig:texture_synthesis}), they result in degraded results as shown in Fig.~\ref{fig:deepfashon}.
In contrast, our human texture synthesis network produces diverse high-quality human textures, effectively reducing the domain gap and improving the performance on real human photos.

\begin{table}
		\begin{center}
			\caption{\textbf{User study of different human textures for sewing pattern reconstruction on real images.} 
   Participants are asked to rate the three methods on a scale of 3 (excellent) to 1 (poor) or 0 (fail in simulation).
   }
			\label{tab:perceptual study}
			\begin{tabular}{lccc}
				\toprule 
				Texture & w/o texture & SURREAL & Ours\\
				\midrule 
				Score $\uparrow$ & 0.822 & 1.12 & \textbf{2.35}\\
				\bottomrule 
			\end{tabular}
		\end{center}
	\end{table}

Furthermore, we conduct a user study for a more comprehensive evaluation.
This study uses 51 images randomly selected from the DeepFashion dataset~\cite{liu2016deepfashion}.
45 subjects are asked to rank the recovered sewing patterns by the models trained using different human textures, with the input image as the reference (1 for poor and 3 for excellent).
We use the averaged scores for measuring the results.
As shown in Table~\ref{tab:perceptual study}, the model trained with the proposed human textures is clearly preferred over other models, suggesting the effectiveness of our algorithm in handling real-world human photos.

\begin{figure}
\includegraphics[width=.7\linewidth]{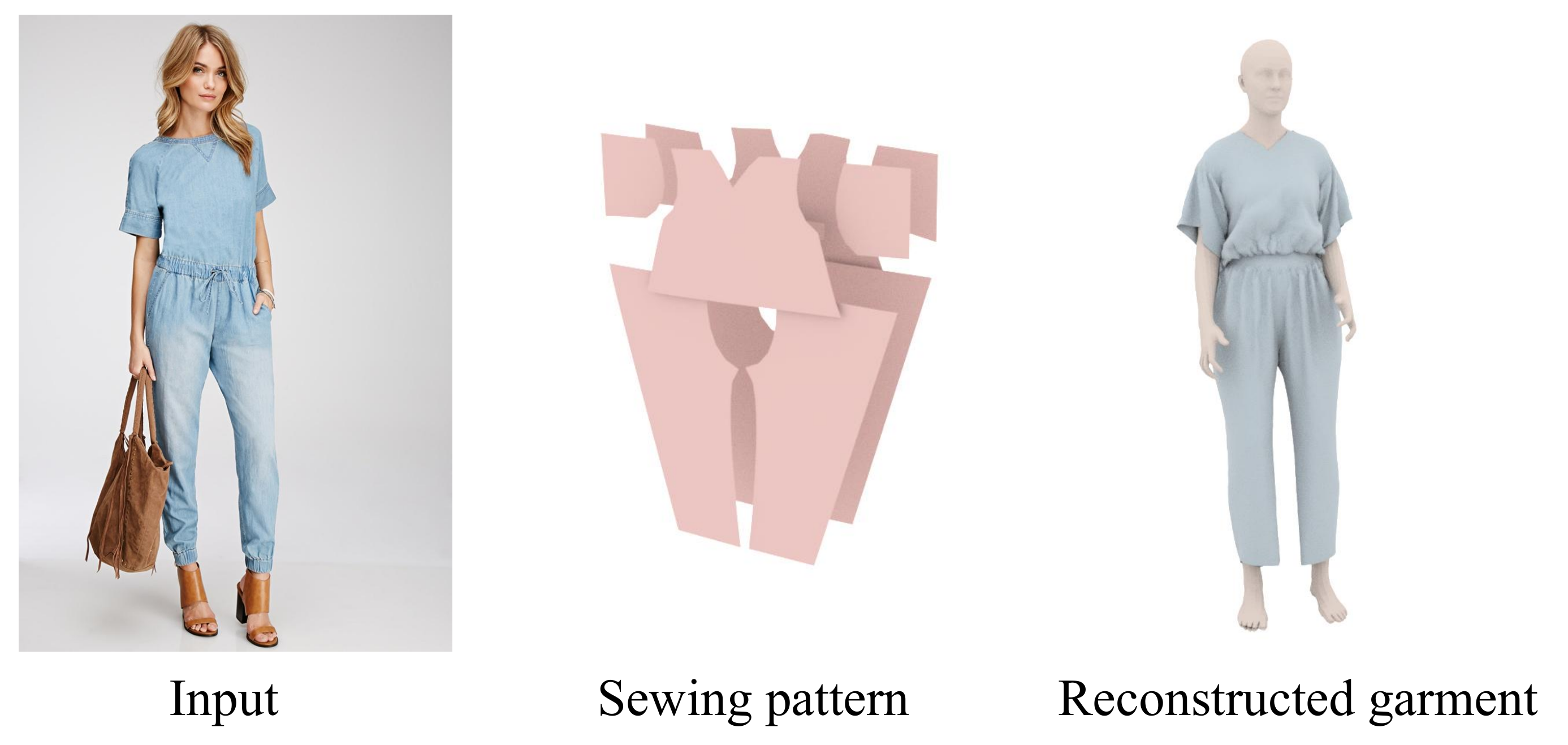}
\caption{
\textbf{Generalization to novel patterns.}
With the template-free network design and the high-quality dataset, the proposed algorithm is able to handle garment topology unseen during training.
}
\Description{figure description}
\label{fig:novel_topology}
\end{figure}
\vspace{1mm}

\paragraph{Generalization to unseen topology.} 
As demonstrated in Fig.\ref{fig:novel_topology}, the proposed algorithm is able to generalize to unseen garment topologies, where the sewing pattern (a one-piece jumpsuit) has not been encountered during training.
This capability can be attributed to two key factors. First, our network is designed without assuming the input garment type. 
This stands in contrast to approaches such as \cite{chen2022structure}, which is constrained by predefined panel groups, or \cite{bhatnagar2019multi}, which relies on fixed garment templates. 
By avoiding these constraints, our network can effectively handle diverse and novel garment topologies.
Second, the SewFactory dataset, specifically designed to prioritize scale and diversity, plays a crucial role in enabling such generalization. 
This comprehensive dataset allows the network to learn the fundamental concept of assembling panels into garments, rather than being restricted to the specific styles present in the training data. 
As a result, our algorithm exhibits the ability to adapt and generate accurate sewing patterns for unseen garment styles.

\begin{figure}
\includegraphics[width=.9\linewidth]{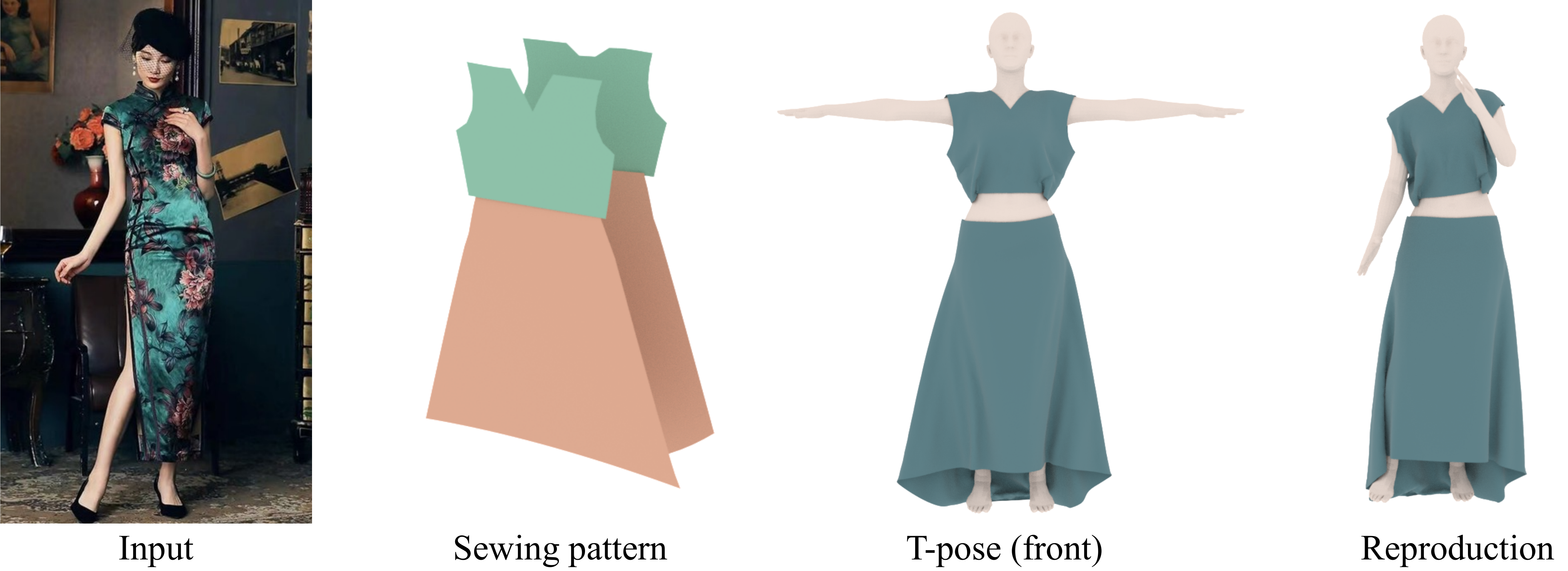}
\caption{
\textbf{Failure case.} The proposed method cannot well handle the classical Chinese dress Qipao whose stitching relations are beyond the settings of our training data. It also lacks the capability to predict unseen accessories such as hats.}
\Description{figure description}
\label{fig:qipao}
\end{figure}

\paragraph{Limitations.} 
While exhibiting strong generalization capabilities to real photos, our method encounters challenges when presented with special inputs that deviate significantly from garments in training data. 
As illustrated in Fig.~\ref{fig:qipao} where we show a classical Chinese dress (known as Qipao), the method may exhibit notable errors in reconstructing stitching relations that are not in training data and lack the capability to predict unseen accessories such as hats. 
To encompass a broader range of clothing styles and accessories can potentially alleviate this issue and improve the model to handle more diverse fashion elements.

 \section{Discussion and Future Work}
 In this work, we tackle the challenging task of recovering garment sewing patterns from a single unconstrained human image. Our contributions include the introduction of the SewFactory dataset, which provides a substantial amount of image-and-sewing-pattern pairs to train data-hungry deep learning models. Additionally, we propose a simple yet powerful Transformer model that achieves high-quality results for sewing pattern reconstruction from a single image.

This work has paved the way for accessible, low-cost, and efficient 3D garment design and manipulation. 
However, there are still several directions for future exploration and improvement.

First, while the proposed Sewformer demonstrates remarkable performance, it is designed in a minimalistic style, and there is still room for enhancing its capabilities. Further investigation could focus on incorporating more advanced attention mechanisms~\cite{liu2021swin} to capture finer details in the sewing patterns, and/or leveraging temporal information from image sequences to enhance the reconstruction. 

Second, expanding the SewFactory dataset and incorporating more diverse garment styles, body shapes and poses, and higher variations in rendering conditions would contribute to better generalization and robustness of the model. 
Furthermore, considering the interaction between garments and human bodies is an intriguing direction for future work. While the proposed SMPL regularization loss shows an effective exploration in this direction, it is possible to take one step further to employ the human body information in a more explicit way, e.g., directly modeling the connection between body parts and panels. 

Lastly, applying the proposed model and dataset in virtual and augmented reality, such as personalized virtual try-on systems, virtual fashion design platforms, or online shopping, could extend the impact of this work into various domains.
	\bibliographystyle{ACM-Reference-Format}
	\bibliography{sample-base}

\end{document}